\PassOptionsToPackage{table}{xcolor}
\documentclass[11pt]{article}

\usepackage[final]{acl}
\usepackage{booktabs} 
\usepackage{times}
\usepackage{latexsym}
\usepackage{amsmath}
\usepackage{mathtools}

\usepackage[T1]{fontenc}
\usepackage{makecell}
\usepackage{amsfonts}

\usepackage[utf8]{inputenc}

\usepackage{microtype}
\usepackage{subcaption}
\usepackage[table]{xcolor}

\usepackage{inconsolata}

\usepackage{graphicx}
\usepackage[most]{tcolorbox}
\usepackage{xcolor}
\usepackage{subcaption}

\definecolor{promptbg}{HTML}{F7F7F7}
\definecolor{promptframe}{HTML}{D0D0D0}

\newtcolorbox{promptbox}{
  colback=promptbg,
  colframe=promptframe,
  boxrule=0.4pt,
  arc=2pt,
  left=4pt,
  right=4pt,
  top=3pt,
  bottom=3pt,
  width=\linewidth,
  fontupper=\small\ttfamily,
  enhanced,
  breakable
}

\title{GraphLit: Learning Text-Enriched Dynamic Character Network Representations for Literary Study}

\author{Gaspard Michel\textsuperscript{\normalfont{1,2}}, Elena V. Epure\textsuperscript{\normalfont{1,3}}, Romain Hennequin\textsuperscript{\normalfont{1}} \AND Christophe Cerisara\textsuperscript{\normalfont{2}},  Mirella Lapata\textsuperscript{\normalfont{4}}
\AND
  \textsuperscript{\normalfont{1}}{\normalfont{Deezer Research, Paris, France}}, 
  \textsuperscript{\normalfont{2}}{\normalfont{Loria, Nancy, France}}, \\
  \textsuperscript{3}{\normalfont{Idiap Research Institute, Switzerland}}, 
  \textsuperscript{4}{\normalfont{School of Informatics, University of Edinburgh}}
  \AND 
  \normalfont{\texttt{\{gaspard.michel, christophe.cerisara\}@loria.fr}}
}

\begin{document}
\maketitle
\begin{abstract}

Methods to represent literary texts as graphs or sequences of graphs mainly focus on representing character interactions, and often overlook another crucial aspect: the textual context in which characters interact.
We introduce Dynamic Heterogeneous Character Networks (DHCNs), which organize long novels into temporally localized heterogeneous graphs that align
characters with their textual contexts.
We extract around 20,000 DHCNs from Project Gutenberg, and propose GraphLit, a self-supervised learning framework that learns rich literary representations through a masked graph autoencoder objective.
Across a wide range of 12 character-related tasks, GraphLit improves over text-only, graph-only and prior hybrid baselines.
Ablations over different kinds of dynamic graph structures and architectural elements show that grounding characters in their context is the main performance driver, while explicitly encoding narrative order and character relationships provide task-dependent improvements.
Finally, we demonstrate the applicability of DHCNs and GraphLit for literary analysis by studying the link between narrative non-linearity and dynamic social features.\footnote{Code is available at \url{https://github.com/gasmichel/GraphLit}.}

\end{abstract}

\section{Introduction}

\begin{figure*}[ht!]
    \centering
     
  \begin{subfigure}[b]{.63\linewidth}
    \includegraphics[width=\linewidth]{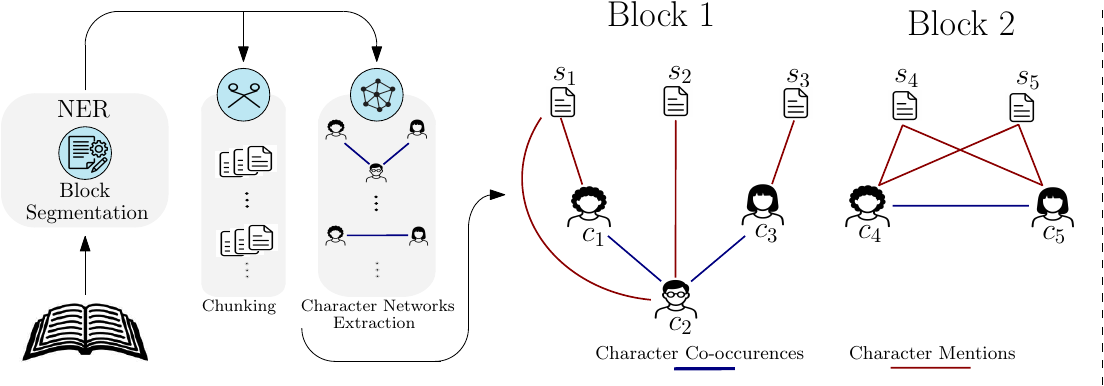}
    \caption{Extraction of \textbf{Dynamic Heterogeneous Character Networks}.}
    \label{fig:extract}
  \end{subfigure}
  \hfill
  \begin{subfigure}[b]{.36\linewidth}
    \includegraphics[width=\linewidth]{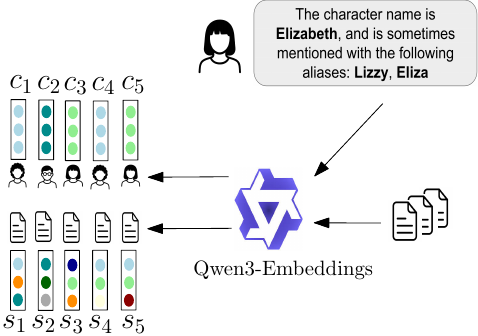}
    \caption{Initial Node Attributes}
    \label{fig:init_attr}
  \end{subfigure}


\caption{DHCN extraction (a) and initialization (b). Novels are divided into blocks and segments; each block yields a heterogeneous graph connecting co-occurring characters to the segments in which they appear. Segment texts and character-alias prompts initialize the two node types.}
    \label{fig:graph}
\end{figure*}

A longstanding debate in literary criticism concerns whether characters should be analysed as narrative functions or as human-like figures with psychological and social dimensions \cite{woloch2003}.
Formalist approaches, such as Propp's, define characters through the roles they fulfil in narrative structures, such as hero or villain \cite{propp1968}.
For Barthes and Greimas, characters are rather ``beings of paper'' that emerge from the text through \textit{signifiers} and only serve a functional purpose in a narrative grammar \cite{Barthes1968, Barthes1975, Greimas1984}.
Zunshine and Palmer emphasize how readers draw cognitive inferences through which they represent characters as social agents 
\cite{Palmer2004, Zunshine2006}.

Computational methods to represent characters mirror this division.
Text-based representations capture descriptions, dialogue, and events \cite{michel-etal-2024-improving, yang-anderson-2024-evaluating, Bourgois2026, zhang-etal-2026-respecting} but do not explicitly represent evolving social structures.
This compromises the ability to explicitly model important aspects such as the prominence in Woloch's \textit{character-space}---or how much narrative space a character is taking with respect to the others \cite{moretti2011network}.
In contrast, \textit{Character Networks} (CNs) explicitly model character relations and their evolution over a narrative  \cite{LabatutNetworks2019} but omit the textual signifiers through which characters are described. 
LLM-based profiles have been explored as a textual alternative to generate rich character descriptions that can capture both relations and signifiers, yet their generalization to unseen novels remains uncertain and their evaluation often relies on copyrighted data that cannot be redistributed \cite{chang-etal-2023-speak, xu-etal-2024-fine, gurung-lapata-2024-chiron, yuan-etal-2024-evaluating}.
Besides, LLMs have been shown to have difficulty understanding important narrative concepts such as ordering \cite{wang-etal-2024-ada} and relationships \cite{hamilton-etal-2026-narrabench},
questioning their reliability for this task.
We therefore seek reproducible representations that combine evolving textual contexts and character relations.

In this work, we introduce a novel structural organization of long novels into temporally localized heterogeneous graphs that align characters with the ordered textual contexts in which they appear, which we call \textbf{Dynamic Heterogeneous Character Networks} (DHCNs).
By explicitly integrating narrative order, character relationships and the textual context in which characters appear, DHCNs provide a hybrid approach for the modeling of novels that combines different theoretical views of fictional characters (Figure~\ref{fig:graph}).
From approximately 20,000 DHCNs extracted from Project Gutenberg novels, we train \textbf{GraphLit}, a self-supervised model that extends the Heterogeneous Graph Masked Autoencoder framework \cite{hgmae} with global character-link prediction and block-ordering objectives. The resulting model learns representations for characters, blocks, and books.

The closest hybrid approach, \citet{inoue-etal-2022-learning}, connects characters to textual signifiers through typed dependency relations \cite{bamman-etal-2014-bayesian} in a static CN.
Instead, DHCNs model full local CNs and their evolving textual contexts over narrative time.
Across twelve character, contextual, and book-level tasks, GraphLit outperforms directly comparable text-only, graph-only, and hybrid baselines, and its character representations substantially improve quotation attribution \cite{michel-etal-2024-improving}.
Ablations show that character-segment grounding drives most downstream gains and that explicit character-character edges provide smaller, but relevant benefits on socially grounded tasks such as role and protagonist prediction.
Additionally, our findings show that temporal encoding has limited impact on tasks defined by static character attributes, but is important for inherent temporal understanding such as narrative order prediction.


We further evaluate GraphLit as a tool for Digital Humanities through a case study of narrative non-linearity, defined as the various types of discrepancies occurring in the narrative discourse, such as a shift across time, space, or character perspectives.
While previous work links non-linearity to semantic discrepancies between consecutive narrative units \cite{toubia2017, PIPER2023101793}, we show that it is also associated with changes in local social structure, and that these associations are captured more strongly by DHCN representations than by text-only representations or DHCNs without character-character edges.

Our contributions are:

\begin{enumerate}
    \item \textbf{DHCNs}, a representation that grounds characters in localized text while retaining optional social and temporal structure.
    \item \textbf{GraphLit}, a self-supervised framework that learns character, block, and book representations through graph reconstruction, global character-link prediction, and narrative-order objectives.
    \item An extensive evaluation showing that textual grounding is the principal source of downstream gains, while social and temporal structure provide task-specific benefits for character modeling and literary analysis.

\end{enumerate}




\section{Related Work}




 
\paragraph{Characters as Social Beings.} Character Networks (CNs) model relationships between characters \cite{LabatutNetworks2019}, and have been extensively applied to literary analysis \cite{moretti2011network}.
Common variants include co-occurrence networks, where characters are connected when they appear in the same short text segment \cite{bonato2016miningmodelingcharacternetworks}, and dialogue networks, where links are derived from conversational interactions \cite{elson-etal-2010-extracting}. 
Graph measures over these networks support literary analysis: for example, \textit{centrality} can identify prominent characters \cite{masias2017}, while high \textit{transitivity} may indicate the presence of multiple social groups \cite{stiller2003a}.

Static networks  are, however, limited in their ability to capture narrative time.
Dynamic networks address this limitation by encoding the order of events, thereby better reflecting the story and character dynamics \cite{agarwal-etal-2012-social, prado2016}, and have been applied to plot understanding \cite{Bost2018} and automatic summarization \cite{bost2019remembering}.
However, both static and dynamic character networks remain limited as they do not encode textual information, restricting their use in representation learning and their ability to associate characters with textual signifiers, such as names or appearances \cite{Barthes1975}.
The work of \citet{iyyer-etal-2016-feuding} presents a text-based approach which extracts spans containing mentions of exactly two characters followed by the modeling of social relationships as trajectories over these pairwise spans.
A similar unsupervised approach is proposed by \citet{chaturvedi2017unsupervised} but uses summaries rather than raw texts.

The closest work to ours is the hybrid approach of \citet{inoue-etal-2022-learning}, which builds a global static heterogeneous graph over a corpus of 17,000 Project Gutenberg Books.
Additional \textit{textual signifier} nodes are built by connecting characters to typed dependency relations \cite{bamman-etal-2014-bayesian}, along with author and book nodes.
Then, character embeddings are built by employing the DeepWalk algorithm \cite{Perozzi_2014} on this large graph.

\paragraph{Characters through Textual Signifiers.} 
Characters could also be represented from the textual context that creates the storyworld.
Then, characters are not only defined by their narrative function or social role; they are also given real or imagined attributes that reinforce readers' social cognition \cite{zunshine2003theory}.
Previous work has modelled characters through latent personas \cite{bamman-etal-2014-bayesian}, dialogues \cite{li-etal-2023-multi-level, michel-etal-2024-improving}, and assertions or events associated with them \cite{inoue-etal-2022-learning}.
Such representations take the form of latent embeddings, which are difficult to interpret from a human perspective.
Thus, recent studies have introduced theory-grounded ontologies that map characters to interpretable narratological classes rather than latent vectors, capturing for example  their \textit{sociological} role  \cite{Bourgois2026} or their \textit{interiority} \cite{mian2026computationalrepresentationscharactersignificance}.

LLMs offer a promising direction by generating textual character profiles on entire novels \cite{yuan-etal-2024-evaluating} or short stories \cite{gurung-lapata-2024-chiron}/
Yet, it remains unclear if LLM-based representations generalize to novels unseen during pre-training \cite{chang-etal-2023-speak}.
Recent work suggests that LLMs struggle to understand important narrative concepts such as ordering \cite{wang-etal-2024-ada} and relationships \cite{hamilton-etal-2026-narrabench},
questioning their reliability for this task.


\paragraph{Graph Representation Learning.} Across fields, this paradigm has proven effective for analysing graph-structured data. 
Despite extensive work on CNs, few studies have applied graph representation learning to literary networks.
\citet{perri2022one} found Graph Neural Networks-based character embeddings from CNs to outperform word2vec representations for character classification in Tolkien's Legendarium. 
\citet{Lee2020} encode substructures of dynamic character networks with Doc2Vec and pool character representations to learn story-level embeddings.
 
Our current work, \textbf{GraphLit}, differs from prior approaches by learning representations on DHCNs. This formulation captures characters through evolving textual contexts over narrative time while incorporating explicit social relations via CN structures, preserving multi-character interactions and higher-order patterns such as triads. GraphLit is trained on a large corpus of publicly available fiction, providing a reproducible framework for learning rich representations. It also extends beyond characters to model broader narrative units, including chapters and books, within a single framework.







\section{Methodology}

Converting novels into DHCNs involves several steps depicted in Figure~\ref{fig:graph}.
First, we apply standard preprocessing including Named Entity Recognition (NER), Character Name Clustering (CNC), and DHCN Extraction \cite{LabatutNetworks2019, Amalvy_2024}.
For CNC, we follow the work of \citet{amalvy-etal-2025-role}.
We chose not to use coreference resolution due to its computational cost and noisy predictions \cite{martinelli-etal-2025-bookcoref}; see Appendix~\ref{app:name_clustering} for further discussion.





\subsection{Notations}
For a book $B$, we define a detected character $c \in \mathcal{C}$ and its associated mention cluster $M_c = \{m_{1,c}, \dots, m_{n, c}\}$, identified by CNC.
In this work, a mention $m_i$ is a named mention of a character such as ``Mr. Bennet'', and $M_c$ represent all of the different aliases referring to character $c$.
We divide each book $B$ into a set of non-overlapping text blocks $\mathcal{T}$
with each block having a fixed size.
For a block $t \in \mathcal{T}$, the Local Character Network (\textbf{LCN}) is defined as $G_t^\mathcal{C} = (V_t^\mathcal{C}, E_t^\mathcal{C})$. 
A vertex $v \in V_t^\mathcal{C}$ corresponds to a character $c \in \mathcal{C}$ if and only if $c$ is mentioned in the block $t$
and an edge $e =(c, c') \in E_t^\mathcal{C}$ is built between characters $c$ and $c'$ if there exists at least two mentions $m_{c}$ and $m_{c'}$ that co-occur in a restrictive window of 20 words, following standard practice in CN extraction \cite{LabatutNetworks2019}.

For an entire book $B$, we then build Dynamic Character Networks (\textbf{DCN}) as the ordered sequence of disconnected LCNs: $(G^\mathcal{C}_t)_{t \in \mathcal{T}}$ \cite{LabatutNetworks2019}.
The same character $c \in \mathcal{C}$ can appear in several LCNs, as shown in Figure~\ref{fig:graph}, and therefore occur multiple times in DCNs.



In the context of representation learning, LCNs and DCNs are limited as they do not explicitly encode the textual contexts in which characters appear.
We thus extend DCNs by adding textual information through the introduction of a new node type: \textit{segment} nodes.
Formally, we split each block $t \in \mathcal{T}$ using a paragraph break into $n$ smaller segments $V^{\mathcal{S}}_{t} = (s_{t, 1}, \dots\, s_{t, n})$.
Based on this augmentation, we add a new link-type: \textit{character-segment}, connecting a \textit{character} $c$ and a \textit{segment} $s_i$, $(v_c, v_{s_i}) \in E^\mathcal{S}_t$ when $c$ is mentioned in $s_i$, where $E^\mathcal{S}_t$ is the set of such edges at block $t$.




We now define the Heterogeneous Character Network (\textbf{HCN}) as $H_t = (\mathcal{V}_t, \mathcal{E}_t, T_\mathcal{V}, T_\mathcal{E}, X^\mathcal{C}_t, X^\mathcal{S}_t)$ where $\mathcal{V}_t = V_t^\mathcal{C} \cup V_t^\mathcal{S}$, $\mathcal{E}_t = E_t^\mathcal{C} \cup E_t^\mathcal{S}$, $T_\mathcal{V}$ is the set of node types (\textit{character} or \textit{segment}), $T_\mathcal{E}$ is the set of edge types (\textit{character-character} or \textit{character-segment}), and $X^\mathcal{C}_t$ and $X^\mathcal{S}_t$ are the multisets of initial attributes for nodes in $V_t^\mathcal{C}$ and $V_t^\mathcal{S}$ respectively.
Graph nodes often carry features that describe the entities they represent. 
In our work, we leverage any textual encoder $\Phi$ to compute initial representations of the textual information contained in \textit{segments} ($X^\mathcal{S}_t$) and \textit{character} ($X^\mathcal{C}_t$) nodes. 
\textit{Segment} nodes are maped directly to embeddings by the textual encoder. 
For \textit{characters}, the only available information is their name and aliases extracted during the CNC step, encoded using the fixed template shown in Figure~\ref{fig:init_attr}.
This allows us to build initial node representations in a comparable embedding space, reducing the potential feature-distribution mismatch common in heterogeneous graphs \cite{hgt2020}.
We use \texttt{Qwen3-4b-Embeddings} \cite{qwen3embedding} as the textual encoder $\Phi(\cdot)$ and a dedicated \textit{task} prompt that frames character representation as a query for retrieving relevant passages (Appendix~\ref{app:losses}).

\begin{figure}[t!]
    \centering
    \includegraphics[width=\linewidth]{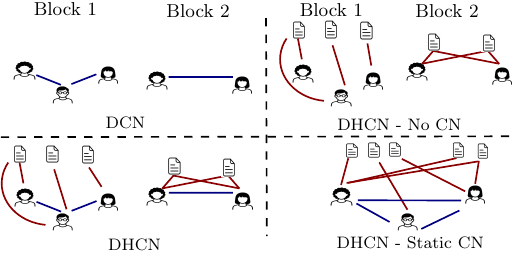}
    \caption{Graph structures evaluated in our experiments: DCN, full DHCN, DHCN without character-character edges (No CN), and DHCN with a static character network (Static CN).}
    \label{fig:types}
\end{figure}

Then, the \textbf{DHCN} for a book $B$ is simply defined as the sequence of disconnected HCNs, $(H_t)_{t\in\mathcal{T}}$.
In this work, we also evaluate DHCNs without \textit{character-character} edges, and DHCNs where the CN structure is static.
In the latter, each character is represented by a single node connected to all segments it was mentioned in and to other  co-occurring character nodes.
Figure~\ref{fig:types} shows the different network types considered in this work.

\subsection{Representation Learning with DHCN}

The GraphLit training framework extends HGMAE \cite{hgmae}, a self-supervised learning framework that attempts to extract information from graph data, incorporating elements tailored for the uniqueness of DHCNs.
During training, HGMAE masks edges and initial node attributes, and learns to reconstruct original graphs from corrupted ones.
At inference, it directly uses the unmasked graph.

\paragraph{Masking Strategy}

Formally, given an input HCN, $H_t$, and its initial edges $\mathcal{E}_t$, we apply a random mask with fixed rate $\tau$ to \textit{character-character} edges $E^\mathcal{C}_t$ and \textit{character-segment} edges $E^\mathcal{S}_t$.
We denote as $\tilde{E}^\mathcal{C}_t$ and $\tilde{E}^\mathcal{S}_t$ the remaining edges that were not masked.
For node features, we sample a masking rate $\delta(i)$ from a linear schedule with respect to epoch $i$, followed by randomly selecting a subset of nodes $\tilde{{V}}^{{type}}_t \subset V^{{type}}_t$ according to the rate $\delta(i)$ for both types of nodes.
We then replace initial representations with:
\begin{equation}
\small
    {\tilde{X}}_v = 
    \begin{cases*}
        [M] & if  $v \in \tilde{V}_t$ \\
        X_v & {if} $v \notin \tilde{V}_t$
    \end{cases*}
\end{equation}
where $[M] \in \mathbb{R}^{d_{qwen}}$ is a learnable mask token shared across both node types.




\paragraph{GraphLit Backbone}

\begin{figure}[t!]
\centering
  \begin{subfigure}[c]{\linewidth}
   \centering
    \includegraphics[width=\linewidth]{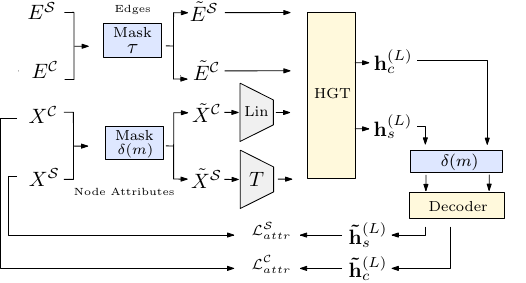}
    \caption{The GraphLit backbone}
    \label{fig:arch_1}
  \end{subfigure}
  \vfill
  \begin{subfigure}[c]{\linewidth}
   \centering
    \includegraphics[width=\linewidth]{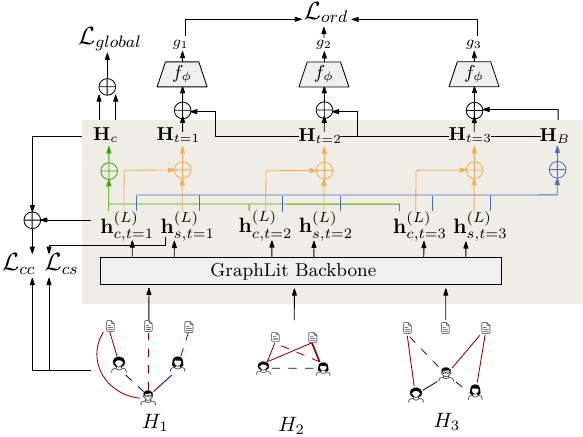}
    \caption{Training GraphLit}
    \label{fig:arch_2}
  \end{subfigure}
\caption{GraphLit architecture and self-supervised training objectives.}
\label{fig:arch}
\end{figure}

We start by projecting (masked) initial node  attributes ($\tilde{X}^\mathcal{C}_t$ and $\tilde{X}^\mathcal{S}_t$), applying different layers, depending on the node type, as shown in Figure~\ref{fig:arch_1}.
For $\tilde{X}^\mathcal{C}_t$, we project textual representations with a linear layer $Lin$.
For $\tilde{X}^\mathcal{S}_t$, the HCN formulation does not explicitly encode segment order due to node permutation invariance. However, preserving the autoregressive nature of text is important for coherence.
To address this, for an HCN at block $t$, we input the ordered sequence of segment attributes $(\tilde{X}_{t,s_1}, \dots \tilde{X}_{t, s_n})$ to a small causal Transformer ($T$) \cite{attention2017}. 
This allows the projected segment embeddings to capture ordering information within their block.

Projected node attributes are then passed to an Heterogeneous Graph Transformer (HGT) backbone with $L$ layers \cite{hgt2020}, that refines node representations $\mathbf{h}^{(l)}_c$ and $\mathbf{h}_s^{(l)}$  by aggregating messages from their neighbours (Appendix~\ref{app:hgt}).
We input a full DHCN $(H_t)_{t=t_1,\dots, t_n}$ to the GraphLit backbone, processing entire novels in a single forward pass.
The final node representations $\{ (\mathbf{h}_{v,t}^{(L)}, \mathbf{h}_{u,t}^{(L)}),  v \in V^\mathcal{C}_t, u \in V^\mathcal{S}_t\}$ are then aggregated to build \textbf{global representations} of characters $c$, blocks $t$ and the entire book $B$:

{\small
\begin{align}
    \hspace{-4pt}\mathbf{H}_c &= \text{POOL}_{\phi_1}(\mathbf{h}_{v,t}^{(L)}, v \in V^\mathcal{C}_t, \; v=c, \; t \in \mathcal{T}) \\
    \hspace{-4pt}\mathbf{H}_t &= \text{POOL}_{\lambda_1}(\mathbf{h}_{v,t}^{(L)}, \mathbf{h}_{u,t}^{(L)}, \; v \in V^\mathcal{C}_t, u \in V^\mathcal{S}_t) \\
    \hspace{-4pt}\mathbf{H}_B &= \text{POOL}_{\lambda_2}(\mathbf{h}_{v,t}^{(L)}, \mathbf{h}_{u,t}^{(L)}, \; v \in V^\mathcal{C}_t, u \in V^\mathcal{S}_t, \; t \in \mathcal{T})
\end{align}
}
where $\text{POOL}_{\phi_1}$ is a multi-head attention layer, and $\text{POOL}_{\lambda_j}$ are heterogeneous multi-head attention layers, described in Appendix~\ref{app:mha_heads}.
We hypothesize that segment and character nodes should contribute differently to global representations.
Attention layers learn to weight the importance of specific character and segment nodes in the final pooled representation.
Figure~\ref{fig:arch_2} summarizes this process.



\paragraph{Block Ordering}

In the DHCN formulation, each HCN in the sequence $(H_t)_{t=t_1,\dots, t_n}$, is a disconnected graph, with no access to information from other blocks.
Similar to the order of segments within a block, the order in which blocks occur in narrative discourse is a crucial signal for story understanding \cite{genette1980narrative}.
While explicit information can be introduced directly in the DHCNs creation, we instead follow a self-supervised paradigm, and choose to incorporate ordering signal through an ordering loss.
Formally, we train a neural network to provide a scalar score for each block representation, augmented with global information from book embeddings $g_t = f_\phi ([\mathbf{H}_t\; | \; \mathbf{H}_B])$
where $f_\phi$ is a Multi Layer Perceptron (MLP).
Then, we maximise the likelihood of the true ordering of $N$ blocks with a ListMLE loss \cite{xia2008listwise}:
\begin{equation}
\mathcal{L_\text{ord}} = -\sum_{t=1}^{N} ( g_{t} - \log \sum_{j=t}^{N} \exp({g_{j})} )
\end{equation}
where~$g_t$ is the score assigned to the block at position~$t$. The objective encourages the model to assign higher scores to earlier blocks, thereby recovering their ground-truth narrative order.

\begin{table*}[t!]
\centering
\small
\renewcommand{\arraystretch}{1}
\setlength{\tabcolsep}{4.25pt}
\begin{tabular}{lccccc|ccccc|cccc}
\toprule
& \multicolumn{5}{c}{\textbf{Character}}
& \multicolumn{5}{c}{\textbf{Context} }
& \multicolumn{4}{c}{\textbf{Book}}  \\

\cmidrule(lr){2-6}
\cmidrule(lr){7-11}
\cmidrule(lr){12-15}
& gender
& role
& prot
& id
& Avg
& spk
& maskd
& desc
& QA
& Avg
& auth
& book
& genre
& Avg 
\\
Support & 4216 & 418 & 4315 & 4000 & --- & 1490 & 893 & 1987 & 345 & --- & 3566 & 3543 & 37332 & --- \\
\midrule 
 \makecell[l]{Hybrid CN\\[-0.1em]{\cite{inoue-etal-2022-learning}}} 
& \makecell[c]{87.2\\[-0.5em]{\tiny (0.1)}}
& \makecell[c]{40.4\\[-0.5em]{\tiny (1.0)}}
& \makecell[c]{69.5\\[-0.5em]{\tiny (0.2)}}
& \makecell[c]{92.8\\[-0.5em]{\tiny (0.1)}}
& \makecell[c]{72.5\\[-0.5em]{\tiny (0.4)}}
& \makecell[c]{54.9\\[-0.5em]{\tiny (0.2)}}
& \makecell[c]{58.5\\[-0.5em]{\tiny (0.5)}}
& \makecell[c]{72.5\\[-0.5em]{\tiny (0.2)}}
& \makecell[c]{49.0\\[-0.5em]{\tiny (0.7)}}
& \makecell[c]{58.7\\[-0.5em]{\tiny (0.4)}}
& \makecell[c]{75.7\\[-0.5em]{\tiny (0.2)}}
& \makecell[c]{88.3\\[-0.5em]{\tiny (0.1)}}
& \makecell[c]{75.4\\[-0.5em]{\tiny (0.2)}}
& \makecell[c]{79.8\\[-0.5em]{\tiny (0.2)}} \\
\midrule
Qwen3-Emb-4b
& \makecell[c]{87.1\\[-0.5em]{\tiny (0.1)}}
& \makecell[c]{34.2\\[-0.5em]{\tiny (0.3)}}
& \makecell[c]{62.1\\[-0.5em]{\tiny (0.4)}}
& \makecell[c]{98.3\\[-0.5em]{\tiny (0.0)}}
& \makecell[c]{70.4\\[-0.5em]{\tiny (0.2)}}
& \makecell[c]{\textbf{56.7}\\[-0.5em]{\tiny (0.2)}}
& \makecell[c]{54.8\\[-0.5em]{\tiny (0.4)}}
& \makecell[c]{72.9\\[-0.5em]{\tiny (0.1)}}
& \makecell[c]{48.7\\[-0.5em]{\tiny (0.5)}}
& \makecell[c]{58.3\\[-0.5em]{\tiny (0.3)}}
& \makecell[c]{72.8\\[-0.5em]{\tiny (0.6)}}
& \makecell[c]{87.3\\[-0.5em]{\tiny (0.3)}}
& \makecell[c]{79.4\\[-0.5em]{\tiny (0.1)}}
& \makecell[c]{79.8\\[-0.5em]{\tiny (0.3)}}
\\
DCN
& \makecell[c]{93.9\\[-0.5em]{\tiny (0.2)}}
& \makecell[c]{41.6\\[-0.5em]{\tiny (0.1)}}
& \makecell[c]{75.8\\[-0.5em]{\tiny (0.4)}}
& \makecell[c]{97.6\\[-0.5em]{\tiny (0.1)}}
& \makecell[c]{77.2\\[-0.5em]{\tiny (0.2)}}
& \makecell[c]{51.4\\[-0.5em]{\tiny (0.4)}}
& \makecell[c]{63.0\\[-0.5em]{\tiny (0.4)}}
& \makecell[c]{68.0\\[-0.5em]{\tiny (0.3)}}
& \makecell[c]{44.4\\[-0.5em]{\tiny (0.7)}}
& \makecell[c]{56.7\\[-0.5em]{\tiny (0.4)}}
& \makecell[c]{72.5\\[-0.5em]{\tiny (0.2)}}
& \makecell[c]{95.3\\[-0.5em]{\tiny (0.1)}}
& \makecell[c]{73.1\\[-0.5em]{\tiny (0.4)}}
& \makecell[c]{80.3\\[-0.5em]{\tiny (0.2)}} \\
No CN
& \makecell[c]{{96.0}\\[-0.5em]{\tiny (0.1)}}
& \makecell[c]{{44.8}\\[-0.5em]{\tiny (0.9)}}
& \makecell[c]{81.7\\[-0.5em]{\tiny (0.2)}}
& \makecell[c]{\textbf{99.5}\\[-0.5em]{\tiny (0.0)}}
& \makecell[c]{{80.5}\\[-0.5em]{\tiny (0.3)}}
& \makecell[c]{52.6\\[-0.5em]{\tiny (0.4)}}
& \makecell[c]{{68.8}\\[-0.5em]{\tiny (0.5)}}
& \makecell[c]{{75.6}\\[-0.5em]{\tiny (0.2)}}
& \makecell[c]{51.6\\[-0.5em]{\tiny (0.5)}}
& \makecell[c]{{62.2}\\[-0.5em]{\tiny (0.4)}}
& \makecell[c]{\textbf{78.3}\\[-0.5em]{\tiny (0.1)}}
& \makecell[c]{{95.6}\\[-0.5em]{\tiny (0.1)}}
& \makecell[c]{\textbf{81.0}\\[-0.5em]{\tiny (0.2)}}
& \makecell[c]{\textbf{85.0}\\[-0.5em]{\tiny (0.1)}} \\
Static CN
& \makecell[c]{\textbf{96.1}\\[-0.5em]{\tiny (0.1)}}
& \makecell[c]{41.8\\[-0.5em]{\tiny (1.0)}}
& \makecell[c]{{82.4}\\[-0.5em]{\tiny (0.1)}}
& \makecell[c]{99.0\\[-0.5em]{\tiny (0.0)}}
& \makecell[c]{79.8\\[-0.5em]{\tiny (0.3)}}
& \makecell[c]{{54.9}\\[-0.5em]{\tiny (0.2)}}
& \makecell[c]{\textbf{71.1}\\[-0.5em]{\tiny (0.3)}}
& \makecell[c]{\textbf{78.7}\\[-0.5em]{\tiny (0.2)}}
& \makecell[c]{\textbf{54.4}\\[-0.5em]{\tiny (0.7)}}
& \makecell[c]{\textbf{64.8}\\[-0.5em]{\tiny (0.3)}}
& \makecell[c]{74.1\\[-0.5em]{\tiny (0.3)}}
& \makecell[c]{\textbf{96.4}\\[-0.5em]{\tiny (0.1)}}
& \makecell[c]{{80.7}\\[-0.5em]{\tiny (0.1)}}
& \makecell[c]{{83.7}\\[-0.5em]{\tiny (0.2)}} \\
 DHCN
& \makecell[c]{95.7\\[-0.5em]{\tiny (0.0)}}
& \makecell[c]{\textbf{48.7}\\[-0.5em]{\tiny (0.4)}}
& \makecell[c]{{\textbf{82.8}}\\[-0.5em]{\tiny (0.2)}}
& \makecell[c]{{99.3}\\[-0.5em]{\tiny (0.0)}}
& \makecell[c]{\textbf{81.6}\\[-0.5em]{\tiny (0.1)}}
& \makecell[c]{52.9\\[-0.5em]{\tiny (0.3)}}
& \makecell[c]{67.0\\[-0.5em]{\tiny (0.4)}}
& \makecell[c]{75.5\\[-0.5em]{\tiny (0.3)}}
& \makecell[c]{{52.1}\\[-0.5em]{\tiny (0.7)}}
& \makecell[c]{61.9\\[-0.5em]{\tiny (0.4)}}
& \makecell[c]{{76.3}\\[-0.5em]{\tiny (0.1)}}
& \makecell[c]{94.6\\[-0.5em]{\tiny (0.1)}}
& \makecell[c]{79.9\\[-0.5em]{\tiny (0.2)}}
& \makecell[c]{83.6\\[-0.5em]{\tiny (0.1)}} \\
\bottomrule
\end{tabular}
\caption{Accuracy (\%) across CEB tasks with varying graph structures. Best results among tested models are highlighted in bold, and standard deviations across model seeds are in parentheses.}
\label{tab:ceb_results}
\end{table*}

\paragraph{Graph Reconstruction}
Following HGMAE, we train GraphLit to reconstruct corrupted DHCNs.
Edge reconstruction is framed as a binary link prediction task 
over masked character-character and character-segment edges with subsampled negatives.
Link prediction is done by augmenting a  local character representation with the corresponding global information: $\mathbf{\hat{h}}_c^{(L)} = \mathbf{h}_c^{(L)} + \mathbf{H}_c$, followed by scoring with a Multi Layer Perceptron (MLP) applied to the concatenation of node embeddings.
We denote as $\mathcal{L}_{cc}$ and $\mathcal{L}_{cs}$ the type-dependent link-prediction losses.
This formulation follows standard definitions in Graph Autoencoders \cite{kipf2016variationalgraphautoencoders}.
For attribute reconstruction, we feed final node representations along with masked edges to a small decoder, set as a single-layer HGT model, and learn to reconstruct initial masked attributes with a scaled cosine loss \cite{hgmae} for both characters ($\mathcal{L}_{attr}^{\mathcal{C}}$) and segments ($\mathcal{L}_{attr}^{\mathcal{S}}$).
Appendix~\ref{app:losses} provides a detailed explanation.

\paragraph{Global Link Prediction}  To build global character embeddings that encode book-level information, we add a global link prediction task.
We sample global positive and negative interactions between characters \textbf{within the same novels} and \textbf{across different novels}, and learn to dissociate characters with an additional negative log-likelihood loss $\mathcal{L}_{global}$ applied to $\mathbf{H}_c$.
The final loss applied to GraphLit is the sum of the ordering, global and standard link prediction and reconstruction losses.

\section{Experiments}

\subsection{Training}

We train GraphLit on DHCNs extracted from a corpus of novels from Project Gutenberg.
We initially use 9,000 books from \citet{kim-skiena-2022-chapter}, to which we add around 11,000 different novels, and use similar validation and test splits.
For CN extraction, we use a co-occurrence window of 20 tokens and remove characters that are mentioned less than 10 times as done in prior work \cite{michel-etal-2024-improving}.
We build DHCNs with segments of approximately 100 tokens that we found superior in terms of ordering performance compared to larger segments, extracted from blocks of 1,500 tokens, which have been proven to be coherent units when analysing narrative non-linearity \cite{PIPER2023101793}.
We also evaluate GraphLit trained on DHCNs where blocks are built at the chapter level (see Appendix~\ref{app:add_results}).
DHCNs' descriptive statistics can be found in Appendix~\ref{app:desc_stats}.

GraphLit uses an HGT backbone with 3 layers each equipped with 8 attention heads and an hidden dimension of 256.
Pooling layers also use 8 attention heads, and the causal Transformer uses 4 Llama decoder layers \cite{grattafiori2024llama3herdmodels} with 4 attention heads and an MLP factor of 2.
 Final representation dimension is  set to 256, resulting in a model with around 8.2 million parameters
We train GraphLit for 50 epochs with a batch size of 64 novels, an edge masking rate of $50\%$ and an attribute masking rate of $50\%$, progressively increasing by intervals of $0.005$ at each epoch.
We use the AdamW optimizer with a learning rate of $0.0005$ and a weight decay of $0.001$
(detailed hyperparameters are in Appendix~\ref{app:training_details}).


\subsection{Evaluation}

\paragraph{Character Embedding Benchmark} We evaluate GraphLit embeddings with linear probing on 11 character-related tasks taken from  the Character Embedding Benchmark (CEB) \cite{inoue-etal-2022-learning}, spanning \textit{character}-, \textit{contextual}- and \textit{book}-level tasks.
We automatically map CEB characters to those extracted during CNC, but some mappings are ambiguous or fail. 
We thus retain only unambiguously identified characters, yielding 60--90\% of data points depending on the task. Appendix~\ref{app:CEB_chars} shows broadly similar mention and alias-count distributions for retained and excluded characters, though this does not rule out differences in referential complexity.
We include two different contextual tasks: character identification from masked character descriptions (\textit{maskd)} \cite{brahman-etal-2021-characters-tell} and \textit{sentence description} extracted from BookWorm \cite{ papoudakis-etal-2024-bookworm}.
For training, we use a setup similar to CEB, which includes k-fold cross validation and allows character evaluation in seen books.
Experimental and dataset details are described in Appendix~\ref{app:CEB}.

\paragraph{Quotation Attribution} As in \cite{michel-etal-2024-improving}, we evaluate character representations on quotation attribution (i.e., find the speaker of an utterance given its context).
We use PDNC \cite{vishnubhotla-etal-2022-project}, a dataset of about 40,000 manually annotated quotations from canonical novels, and manually map its characters to DHCN characters.
We use all character mentions extracted from DHCNs within a 200-token context window as candidates and fine-tune ModernBERT \citep{warner-etal-2025-utilizing} for quotation attribution by concatenating contextual mention representations with character embeddings. 
Details are provided in Appendix~\ref{app:add_results}.


\begin{figure}[t!]
    \centering
    \includegraphics[width=1\linewidth]{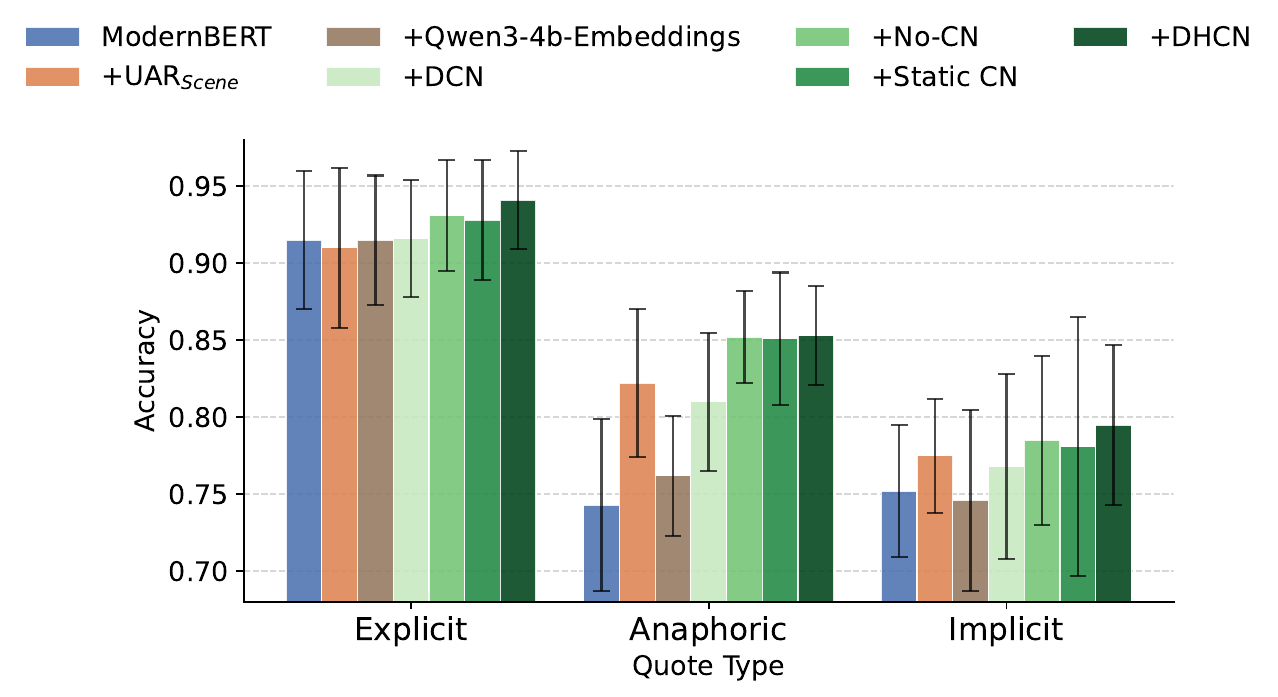}
    \caption{Cross-validation mean accuracy on PDNC by quotation type, along with standard deviation across splits. Models other than ModernBERT (blue) are fine-tuned ModernBERT models which also use character embeddings to form predictions.}
    \label{fig:speaker_attr}
\end{figure}

\paragraph{Baselines} We compare against our reimplementation of the hybrid approach of \citet{inoue-etal-2022-learning} (details in Appendix~\ref{app:CEB}), which uses exactly the same set of characters extracted from our CNC step, thus ensuring fair comparison across methods.
We also include a text-only baseline that averages all \texttt{Qwen3-4b-Embeddings} from segments $s$ in which $c$ was mentioned, and a graph-only baseline that trains a GraphLit variant on DCNs.
We also include GraphLit trained on DHCNs variants: graphs where \textit{character-character} edges are removed (No-CN, Figure~\ref{fig:types}, \textit{top-right}) and graphs where the dynamic character network is replaced by a static network (Static CN, Figure~\ref{fig:types}, \textit{bottom-right}).
All GraphLit baselines have approximately the same number of parameters.
For quotation attribution, we compare against a fine-tuned ModernBERT \cite{warner-etal-2025-smarter} and the same model that uses stylistic character embeddings derived from quotations (UAR$_{Scene}$, \citealt{michel-etal-2024-improving}).

\subsection{Results}
\label{sec:results}


Table~\ref{tab:ceb_results} shows that DHCN representations outperform the directly comparable text-only (Qwen3-Emb-4b), graph-only (DCN), and Hybrid CN baselines, with clear gains on character-level, description, and QA tasks.
We discuss our findings below.


\paragraph{Character-segment grounding is the main performance driver on CEB.} Surprisingly, our ablations show that GraphLit trained with removed character interactions (No CN) is competitive against DHCNs, reaching slightly higher average accuracy on Book and Context tasks.
This is also true for quotation attribution (Figure~\ref{fig:speaker_attr}): only small accuracy differences are observed.
Thus, the dominant information learned by GraphLit appears not to come from character interactions, but rather from the grounding of characters in their dynamic textual contexts.
This counter-intuitive result echoes the debate among literary critics, where our proposed framework focuses rather on the textual signifiers of characters than a story's social structure.

\begin{figure}[t!]
    \centering
    \includegraphics[width=\linewidth]{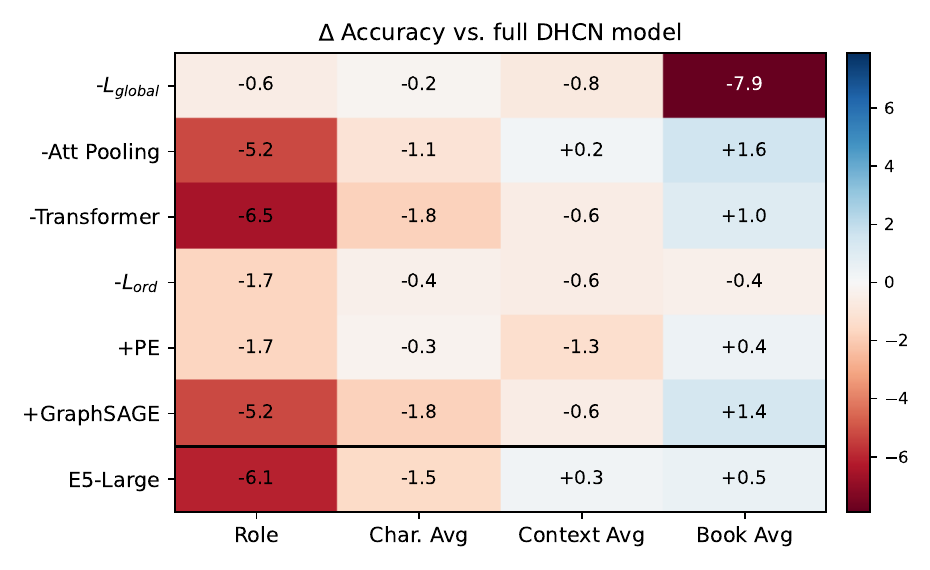}
    \caption{Variation in accuracy points on CEB for GraphLit ablations. The baseline uses DHCNs. We report scores on the role task and task-type averages.}
    \label{fig:ablations}
\end{figure}

\paragraph{CN structure matters for socially grounded tasks.} 
On the \textit{role} and \textit{protagonist} task, all graph methods largely improve over Qwen3-Emb-4b and Hybrid CN, indicating that tested network structures are relevant for socially grounded tasks which require understanding relationships among characters.
Interestingly, even when character networks are removed, No CN performs better than DCN, which suggests that the character-segment grounding allows the network to still learn social patterns.
However, GraphLit trained with DHCNs shows the largest performance on \textit{role} and \textit{protagonist}, suggesting that the explicit modeling of dynamic character relationships is still relevant for socially grounded tasks.
We further evaluate the importance of social grounding through additional quantitative and qualitative analyses in Appendix~\ref{app:social_grounding}.


\paragraph{GraphLit embeddings help with attributing character quotations.}
As shown in Figure~\ref{fig:speaker_attr}, when using GraphLit embeddings with heterogeneous graph structures, we observe large gains in the attribution of \textit{anaphoric} and \textit{implicit} quotations against our adapted ModernBERT baseline, reaching a maximum of +11 and +4 accuracy points respectively with DHCNs.
These experiments directly add character information to the adapted baseline and thus ensure direct comparison.
Such absolute gains are substantial: \citet{michel-etal-2024-improving} reported improvements of around 2 and 3 points against a similar baseline.
In addition, GraphLit embeddings perform substantially better than stylistic embeddings (UAR$_{Scene}$), which are designed to directly improve attribution accuracy (overall accuracy of $0.865$ against $0.837$ respectively).
These results highlight the capability of GraphLit at difficult contextual inference tasks.

\subsection{Architectural Ablations}
\label{sec:ablations}

We analyse the impact of removing or replacing GraphLit's architectural elements to better grasp how they affect its performance.
These ablations include (1) removing the global character loss $L_{global}$ during training, (2) replacing the attention pooling modules $\text{POOL}_{\phi_1}$ and $\text{POOL}_{\lambda_j}$ with simple mean pooling, (3) replacing the causal Transformer that refines segment embeddings with a simple linear layer, (4) removing the ordering loss $L_{ord}$, (5) replacing global narrative order information ($L_{ord}$) with standard Positional Embeddings added to input node embeddings before being fed to the GNN backbone, (6) replacing the heterogeneous GNN backbone with an homogeneous GNN backbone (we use GraphSAGE, \citealt{graphsage}) and (7) replacing the \texttt{Qwen3-4b-Embeddings} encoder backbone with \texttt{multilingual-e5-large} \cite{wang2024multilinguale5textembeddings}.
Ablations (3)-(5) evaluate the impact of modeling different temporal information (within-chunk and global narrative order) while (6) evaluates the necessity of the heterogeneous formulation and (7) the impact of replacing initial node attributes encoded by a non-LLM-based encoder.

Results for the role task and per-task-type averages are reported in Figure~\ref{fig:ablations}, while full per-task results are in Appendix~\ref{app:add_results}.
We see that \textbf{the global character loss is essential for book-level tasks}  (-7.9 points), with negligible effect elsewhere.
$L_{global}$, designed to distinguish characters across books, explains the large gains on cross-book/cross-author tasks. It is thus a core GraphLit component.

We also see that \textbf{Attention Pooling better encodes CN structure}: 
it falls short on the role prediction task (-5.2 points), explaining the lower average on character-level tasks.
Role prediction is where social grounding is relevant, suggesting that attention pooling learns to capture information from character nodes that have crucial social structures.
\textbf{We observe a similar pattern when removing the causal Transformer} (-6.5 points) and using an \textbf{homogeneous graph structure} with GraphSAGE (-5.2 points).
Encoding within-chunk temporal information and explicitly differentiating character from segment nodes through our heterogeneous formulation impacts social grounding.

\begin{table}[t!]
\centering
\small
\renewcommand{\arraystretch}{0.9}
\renewcommand{\cellalign}{ll}
\setlength{\tabcolsep}{4pt}
\begin{tabular}{lccc}
\toprule
 & \textbf{DHCN} & \textbf{No CN} & \textbf{Qwen} \\
\midrule

\textbf{New Chars} \\
\;\;\;\;\; AC 
    & $-0.519^{\star\star}$ & $-0.417^{\star\star}$ & $-0.362^{\star\star}$ \\

\;\;\;\;\; MEAN & $4.485^{\star\star}$ & $3.029^{\star\star}$ & $3.552^{\star\star}$ \\

\textbf{Transitivity} \\ 
\;\;\;\;\; AC
   & $-0.327^{\star\star}$ & $-0.225^{\star\star}$ & $-0.114$ \\

\;\;\;\;\; MEAN
    & $1.655^{\star\star}$ & $0.552^{\star\star}$ & $2.294^{\star\star}$ \\

\textbf{Components} \\ 
\;\;\;\;\; AC
     & $-0.277^{\star\star}$ & $-0.265^{\star\star}$ &  $-0.2615^{\star\star}$ \\

\;\;\;\;\; MEAN
& $0.009$ & $0.023$ & $0.2746^{\star\star}$ \\

\midrule

$R^{2}$
    & 0.322 & 0.441 & 0.2515 \\

\bottomrule
\end{tabular}
\caption{Regression results per type of log-circuitousness ($^{\star\star}$: $p < 0.01$). AC indicates autocorrelation, and MEAN represents average values.
Wald tests show that New Chars AC and Transitivity AC are statistically different between DHCN and other types.}
\label{tab:reg}
\end{table}

Replacing the initial node attributes with E5-large substantially degrades performance on the role task, but slightly improves the context and book tasks. 
GraphLit trained with more capable encoders appears to better capture social structure, but not necessarily character-segment grounding. 
This highlights GraphLit’s versatility: less capable encoders may improve efficiency without reducing performance. 
These results also suggest that potential book memorization by Qwen3 does not fully explain GraphLit’s performance.

Ablations related to global narrative order highlight an interesting observation.
Either completely removing $L_{ord}$ or replacing it with simple Positional Embeddings only slightly degrades overall performance on contextual and character-related tasks.
This suggests that \textbf{the encoding of temporal information is not necessary for GraphLit to learn embeddings that achieve strong performance on CEB}.
This contradicts standard narrative theory, such as Genette's concept of \textit{order} \cite{genette1980narrative},  which emphasises the importance of the ordering of story elements when analyzing narrative discourse.
We hypothesize that this discrepancy between theory and computational evaluation is rather an artefact of CEB, where characters are modeled through fixed-attributes (e.g., the ``archduchess'' role) that do not incorporate story evolution (the archduchess might lose her title).
Building character understanding benchmarks that represent knowledge of characters at fixed discourse times is an exciting direction for future work, allowing us to evaluate the impact of temporal information.

\section{Detection of Narrative Non-Linearity}

Non-linearity, a discrepancy that occurs in a story when the next logical step is replaced by a shift of focus in a different direction, is a notion that has always been present in human narration.
Recently, \citet{toubia2017} proposed \textit{circuitousness}, an embedding-based measure that captures non-linearity when used with Word2Vec embeddings \cite{PIPER2023101793}.
Highly circuitous books are narratives that feature an ordering of events that does not match the expected order of events.

In our analysis, we project each block with \texttt{Qwen3-4b-Embeddings}, and contrast the analysis by comparing circuitousness calculated with GraphLit block embeddings $\mathbf{H}_t$, in the full DHCN and No CN formulations.
To analyse the extent to which narrative non-linearity is also a social phenomenon, we compute \textit{topological features} from local CNs which characterize social relationships among characters:
\textit{transitivity}, \textit{connected components} and \textit{ratio of new characters}, and see their impact on circuitousness.
A high \textit{transitivity} and number of \textit{connected components} imply the presence of separated social groups, while a high \textit{ratio of new characters} from one block to another implies a change of character focalization in the narrative.
Given the sequence of character-networks $G_t^\mathcal{C}$, we compute the associated sequence of features, followed by calculating two book-level statistics: 1) feature values averaged over chunks and 2) feature autocorrelation at lag 1, which informs how each feature is evolving from one block to another, thus providing a global measure of social discrepancy over time.
We provide extra details in Appendix~\ref{app:circuit}.



Following \citet{PIPER2023101793}, we analyse the impact of our measures of social discrepancy on circuitousness using a linear regression, where we control for text length, number of characters, and a dummy variable that takes a value of 1 if the novel was seen at training time.
We use a set of 6084 books from which we gather genre information using metadata from Project Gutenberg, and ensure that no books are tagged as Romance because of their stylistic outlier status 
\cite{wilkens2016genre}.
We standardize log-circuitousness across books, and use it as the dependent variable in the regression.

We present regression results in Table~\ref{tab:reg} and discuss per-genre results in Appendix~\ref{app:circuit}.
Most of the social measures significantly impact circuitousness, with the exception of the average number of components.
Stories with higher average social transitivity or larger average alternation between characters tend to be more circuitous.
In contrast, when social patterns repeat consistently across story blocks (high AC), the narrative tends to be less circuitous.
The regression coefficients indicate that GraphLit embeddings trained with DHCN account more for social measures than Qwen embeddings and DHCN without CN structure, especially in the case of autocorrelations, which measure social discrepancy.
Thus, different embeddings shows different pictures: while Qwen embeddings favour semantic (dis)similarities, GraphLits' embeddings integrate social discrepancies at a higher rate.
This is corroborated when looking at the most circuitous books (DHCNs): these are often compilations of works, featuring distinct plot-lines and social structures.



\section{Conclusion}

We introduced DHCNs and GraphLit, a self-supervised framework for learning character, block, and book representations from localized textual context and dynamic social structure.
Across CEB and quotation attribution, GraphLit representations yield large improvements over comparable graph-only, text-only, and hybrid baselines.
Ablations over graph structure and architectural elements show that textual grounding drives most gains, while explicit character relations provide task-specific benefits.
Our narrative non-linearity case study further demonstrates GraphLit’s value for computational literary analysis.


\section{Limitations}

Our work bears several limitations.

\paragraph{Reliance on noisy CNC}
As with all character network methods, DHCNs identify and group mentions of characters through an imperfect Character Name Clustering (CNC) step.
Thus, for some novels, DHCNs might fail to link crucial mentions of characters together, which can hinder downstream analysis on some characters.
While coreference resolution is often used in the extraction, we chose not to use it because it is a noisy and computationally intensive process.
However, we note that extending GraphLit to coreference-based character networks is a promising direction for future work.

\paragraph{Graph Extraction Hyperparameters}
Our graph extraction process relies on important hyperparameters that affect the resulting graph structure: the co-occurence window and minimal number of mentions for character-network extraction, and the block size and segment size for textual information.
While we did look at how block size impacts downstream performance, we think that measuring the effect of a change in these parameters in both the final structures as well as downstream performance is an interesting future work.


\paragraph{Evaluation Scope}
While CEB provides various \textit{character}-related tasks, it does not encompass how character representations can help on broader downstream tasks, such as description generation, entity linking or long-document understanding.
While we use quotation attribution as an example of such downstream tasks, it would be beneficial to extend the analysis to other downstream tasks.
We note that the applicability of character representations working jointly with Large Language Models is a promising direction for generalization to unseen tasks.

\paragraph{Qwen3-based attributes}
Because Qwen3 may have encountered Project Gutenberg during pretraining, its initial attributes may be affected by memorization. Results with E5-Large provide a non-LLM control with comparable aggregate performance, but do not establish contamination-free generalization to unseen texts.

\paragraph{Reliance on Quantitative Literary Study}
Our case-study on narrative non-linearity mainly includes a quantitative analysis of correlation between DHCNs' topological measures and circuitousness.
Although our analysis is extended in Appendix~\ref{app:circuit} (for example, we discuss that GraphLit's most circuitous books are either volumes or compilations of different works, which are by nature non-linear), a complete qualitative study done with literary scholars would be a promising direction for future work.

\section{Acknowledgements}
We thank all reviewers for their constructive feedback during the rebuttal phase. 
This work was performed using HPC resources from GENCI–IDRIS (Grant AD011017601). Lapata gratefully acknowledges the support of the UK Engineering and Physical Sciences Research Council (grant EP/W002876/1).





\bibliography{anthology_aa,anthology_ab,custom, bilbio}
\appendix

\section{Character Name Clustering}
\label{app:name_clustering}

Following standard procedures of character network extraction \cite{LabatutNetworks2019}, we first use NER on the entire book and only keep mentions that are tagged as character mentions (\texttt{PER} tag).
Then, we follow \citet{amalvy-etal-2025-role} and perform Character Name Clustering (CNC) to regroup \texttt{PER} mentions referring to the same character together.
Formally, for each book, we define a graph where each vertex is a character mention inferred by NER, followed by applying a set of inclusion and restriction heuristics on connections between vertices.
We chose not to use coreference resolution as additional clustering information, namely because coreference resolution on full books is a noisy and computationally intensive process \cite{martinelli-etal-2025-bookcoref}.
We note that perfectly identifying all characters of a fictional work is a very challenging task that has been investigated in prior works \cite{vala-etal-2015-mr, amalvy-etal-2025-role}, and it is thus expected that discovered mention clusters might not fully reflect the reality of all characters.

The full CNC process is described below.
For each vertex in the graph, we infer its gender based on a priority list: 1) if the mention contains a title, we infer the gender based on it, 2) we use a list of around 8000 common English names with their associated gender, and assign a gender to a name if it is non-ambiguous\footnote{\url{https://www.cs.cmu.edu/Groups/AI/areas/nlp/corpora/names/}} and 3) we calculate the occurrence of gendered pronouns (\textit{his}, \textit{himself, her, herself}) situated in a window of 3 words around each occurrence of mentions, and assign a gender based on a majority vote.
Then, we add an edge between vertices following a set of rules based on name variations, and additionally use a gazetteer of around 1000 names and their variations (\textit{e.g.} Tim and Timmy) to link vertices.
Finally, we use heuristics to remove unlikely connections: 1) if the two names have a different inferred gender, 2) if the two names share a common surname but have a different first name and 3) if the two names contain a different honorific (\textit{e.g.} ``Mister'' and ``Mr.'').
The standard procedure then proceeds to remove all edges belonging to the shortest path between prohibited pairs \cite{vala-etal-2015-mr}.
 We also include an additional heuristic that removes isolated vertices (\textit{i.e.} mentions that are not connected to any other vertices) in cases where these isolated vertices can be attributed to multiple connected clusters.
This heuristic targets the removal of isolated last names that occur when two characters have the same last name and an inferred mention is this sole last name (\textit{e.g.} ``Bennet'' which could be attributed to both ``Mr. Bennet'' and ``Elizabeth Bennet'').


\section{Background on Heterogeneous Graph Transformer}
\label{app:hgt}

As most Graph Neural Networks (GNNs), Heterogeneous Graph Transformer (HGT) \cite{hgt2020} processes graph 
data through a cascade of Message Passing layers, which are neural networks that aggregate neighbouring information to update a node representation.
Compared to standard GNNs, HGT processes heterogeneous graphs $H = (V, E, T, X)$ where $V$ is the set of nodes, $E$ the set of edges, $T$ the set of relation types for all edges and $X$ are initial attributes.
As we have shown, Heterogeneous Character Networks only support two types of nodes (\textit{character} and \textit{segments)} as well as two types of relations (\textit{character-character} and \textit{character-segment}).
However, HGT operationalises attention in the general framework of heterogeneous graphs.
In particular, it introduces multi-head Heterogeneous Mutual Attention to capture the difference between distribution types of nodes and edges, when aggregating neighbouring information.
Formally, let $e =(u,v)$ be any edge connecting any node $u$ to any node $v$. Then, an attention score is calculated for all edges with a multi-head attention layer that accounts for 1) the type of the source node $u$, 2) the type of the target node $v$ and 3) the type of the edge $(u,v)$.
These attention scores are then combined with a \textit{message} from a source node $u$ to a target node $v$, that also accounts for different types.
Then, for a target node $v$, its updated representation is calculated as the aggregation of all messages from its neighbours $u \in \mathcal{N}{(v)}$ weighted by the associated attention scores.


\section{Homogeneous and Heterogeneous Multi-Head Attention}
\label{app:mha_heads}

We define below our implementation of Homogeneous and Heterogeneous multi-head attention, used to derive pooled global representations of characters (homogeneous), blocks, and books (heterogeneous).
Suppose the following DHCN: $(H_t)_{t\in\mathcal{T}}$ where $H_t=(\mathcal{V}_t, \mathcal{E}_t, T_\mathcal{V}, T_\mathcal{E}, X_t^\mathcal{C} X_t^\mathcal{S})$, and let $\mathbf{h}^{(L)}_c \in \mathbb{R}^d$ and $\mathbf{h}^{(L)}_s \in \mathbb{R}^d$ be final character and segment representations respectively, let $\tau(u): \bigcap_t \mathcal{V}_t^\mathcal{C}\xrightarrow[]{}\mathcal{C}$ be a function that returns the character identifier $c$ of any character node $u$ in the DHCN.
Homogeneous multi-head attention follows the multi-head attention of \citet{attention2017} with $Q=K=V=\mathbf{h}^{(L)}_u$, where the softmax is taken over the set $I(c) = \{u, 
u \in  \bigcap_t \mathcal{V}_t^\mathcal{C}, \; \tau(u) = c\}$.
For heterogeneous multi-head attention, let $I(t) = \{u,  u \in  \mathcal{V}_t\}$ and $I(B) = \bigcup_{t\in\mathcal{T}} \mathcal{V}_t$ be the set of all nodes in block $t$ and the set of all nodes in book $B$, respectively.
We define the following heterogeneous multi-head attention layer:

\begin{align*}
    m_s^i &= \frac{(\mathbf{W}_{Q_s}^i \mathbf{h}^{(L)}_s)(W_{K_s}^i \mathbf{h}^{(L)}_s)}{\sqrt{d_h}}, \; s \in \mathcal{V}^\mathcal{S} \\
    V_s^i &=  \mathbf{W}_{V_s}^i \mathbf{h}^{(L)}_s \\
    m_c^i &= \frac{(\mathbf{W}_{Q_c}^i \mathbf{h}^{(L)}_c)(W_{K_c}^i \mathbf{h}^{(L)}_c)}{\sqrt{d_h}}, \; c \in \mathcal{V}^\mathcal{C} \\
     V_c^i &=  \mathbf{W}_{V_c}^i \mathbf{h}^{(L)}_c \\
     V^i &= [V_c || V_s], \;\; \alpha^i = \underset{c,s 
    \in I}{\text{softmax}}([\alpha_c||\alpha_i]) \\
    o &= [\underset{i=1,\dots,h}{||}{\alpha^i V^i}]W_O
\end{align*}

where $||$ denotes concatenation, $\mathbf{W}_{Q_s}^i, \mathbf{W}_{Q_c}^i, \mathbf{W}_{K_s}^i, \mathbf{W}_{K_c}^i, \mathbf{W}_{V_s}^i, \mathbf{W}_{V_c}^i \in \mathbb{R}^{d_h \times d_{hgt}} $ are type-specific query, key and value projection layers, $h$ is the number of attention heads, $\mathbf{W}_{O} \in \mathbb{R}^{(d_h . h)\times d_{out}}$ is the output projection, and $I \in \{ I(t), I(B)\}$ is the node set that determines whether we're computing global block or book representations.
Concretely, this layer applies type-specific $Q,K,V$ projections, which are then concatenated before applying softmax normalization to compute attention weights.

\begin{figure}[t]
  \centering
  \begin{minipage}{0.75\linewidth}
    \begin{promptbox}
Instruct: Given a query that contains a character name and its aliases, retrieve book passages relevant to the query

Query: The name of the character is \{name\}, and is sometimes mentioned with one of the following aliases: \{aliases\}
    \end{promptbox}
  \end{minipage}
  \caption{Template used to derive initial character attributes. We use the aliases extracted during character name clustering.}
  \label{fig:prompt_qwen}
\end{figure}

\section{Edge and Attribute Reconstruction}
\label{app:losses}

\paragraph{Initial Node Attributes} are computed using \texttt{Qwen3-4b-Embeddings}.
While segment texts are directly encoded, character prompts are created with a template that uses a task prompt and extracted aliases for each character.
The prompt is provided in Figure~\ref{fig:prompt_qwen}.

\paragraph{Edge Reconstruction} For both edge types, we train a different MLP on the task of link prediction, projecting the concatenation of final node representations to provide a scalar score $l(u,v) = \text{MLP}([\mathbf{h}_v^{(L)} | \mathbf{h}_u^{(L)}])$.
At this stage, we \textit{augment} character representations $\mathbf{\hat{h}}_u^{(L)} = \mathbf{h}_u^{(L)} + \mathbf{H}_{\tau(u)}$ by adding their global representation computed from the entire novel.
This process helps build global embeddings $\mathbf{H}_c$ that add information on local tasks, such as link prediction.
Let $E_t^{\mathcal{C}+}$ denote the \textit{character-character} edges that were initially masked and $E_t^{\mathcal{C}-}$ subsampled negative edges that do not exist in the initial graph $H_t$ (we use similar notations for $E_t^\mathcal{S}$).
We also denote as $\mathcal{G}^+$ a set of global character interactions that occur in the book and $\mathcal{G}^-$ as sampled interactions within the book and across different books that did not occur.
For example, consider the edge $(u,v)$, $u,v \in \mathcal{V}^\mathcal{C}$, then $(\tau(u), \tau(v))$ might be part of $\mathcal{G}^+$.
In particular, we build $\mathcal{G}^-$ by sampling negative edges both from within the same book and across different books.
The link prediction tasks that we use in GraphLit are defined as follows:

{\small
\begin{align*}
    \text{NLL}(x, E^+, E^-) &= \sum_{i\in E^+}\log \sigma(x_i) + \sum_{j \in E^-}\log(1-\sigma(x_j)) \\
    \mathcal{L}_{cc} &= \text{NLL}\left( f_\mathcal{C}([\mathbf{\hat{h}}_{c}^{(L)}||\mathbf{\hat{h}}_{c'}^{(L)}]),  
    \; E_t^{\mathcal{C}+}, E_t^{\mathcal{C}-}\right) \\
    \mathcal{L}_{cs} &= \text{NLL}\left(f_\mathcal{S}([\mathbf{\hat{h}}_{c}^{(L)}||\mathbf{{h}}_{s}^{(L)}]), E_t^{\mathcal{S}+}, E_t^{\mathcal{S}-}\right) \\
    \mathcal{L}_{global} & = \text{NLL}\left(f_\mathcal{G}([\mathbf{H}_c || \mathbf{H_{c'}}]), G_t^+, G_t^- \right)
\end{align*}
}%
where $||$ denotes concatenation, $f_\mathcal{C}, f_\mathcal{S}, f_\mathcal{G}$ are MLP layers that produce scalar scores.

\paragraph{Node Attribute Reconstruction} Following HGMAE \cite{hgmae}, we input the (masked) final node representations $\mathbf{{h}}_{c}^{(L)}$ and $\mathbf{{h}}_{s}^{(L)}$ to a small decoder, which is defined as a different HGT instance with 1 layer, with output hidden dimension set to the initial representation size $d^{dec}$ = $d^{qwen}$.
Before feeding them to the decoder, we leverage again the initial mask $\delta(m)$ to mask again final node representations at the initial sampled positions $\tilde{V}^\mathcal{C}$ and $\tilde{V}^\mathcal{S}$, with a new learnable mask token.
Denoting the masked initial attributes as $\tilde{X}$ and the output of the decoder as $\mathbf{{\tilde{h}}}_{c}^{(L)} \in \mathbb{R}^{d^{dec}}$ and $\mathbf{{{\tilde{h}}}}_{s}^{(L)} \in \mathbb{R}^{d^{dec}}$ we compute the following cosine losses, scaled by a factor of $\gamma$:
\begin{align*}
    \mathcal{L}_{attr}^{C} &= \frac{1}{|\tilde{V}^\mathcal{C}|}\sum\limits_{c \in \tilde{V}^\mathcal{C}}\left( 1 - \frac{\tilde{X}_c \cdot \mathbf{{\tilde{h}}}_{c}^{(L)}}{|| \tilde{X}_c || \times ||\mathbf{{\tilde{h}}}_{c}^{(L)}||}\right)^\gamma \\
    \mathcal{L}_{attr}^{S} &= \frac{1}{|\tilde{V}^\mathcal{S}|}\sum\limits_{s \in \tilde{V}^\mathcal{S}}\left( 1 - \frac{\tilde{X}_s \cdot \mathbf{{\tilde{h}}}_{s}^{(L)}}{|| \tilde{X}_s || \times ||\mathbf{{\tilde{h}}}_{s}^{(L)}||}\right)^\gamma
\end{align*}
In our experiments, we use $\gamma=1$.

\paragraph{Complete Loss} The final loss formulation for GraphLit is the following: 

\begin{align*}
\mathcal{L} = &\lambda_1\mathcal{L}_{ord} + \lambda_2\mathcal{L}_{cc} + \lambda_3\mathcal{L}_{cs} + \lambda_4\mathcal{L}_{global} + \\ &\lambda_5\mathcal{L}_{attr}^\mathcal{C} + \lambda_6\mathcal{L}_{attr}^\mathcal{S}
\end{align*}

In our experiments, we set $\lambda_1 =0.2$ to balance the ordering signal, and use $\lambda_i=0.5$ for $i=2,\dots,6$.

\section{DHCNs Statistics}
\label{app:desc_stats}

We display in Table~\ref{tab:desc_stats} the descriptive statistics of the DHCNs extracted from our dataset.
When extracting blocks of 1500 tokens, we leverage 20,413 books, from which 16,367 are used for training, 710 for evaluation and 1776 for the testing of downstream narrative order.
The remaining approximately 1,500 books were used as test books only in the CEB benchmark, and were not seen during training.
When using chapters as separators, we kept 12,563 books for training, using similar validation and test sets.
In this case, we removed from training data books that contain: 1) a too large amount of chapters (60 chapters), 2) a block with a too large amount of text chunks (150), 3) fewer than 5 total number of character-character edges and 4) no character nodes and 5) books that contain fewer than 5 chapters.
For CEB evaluation, we do not apply any restrictions and instead compute embeddings on all DHCNs.
We see that DHCNs in the chapter split contain slightly more character nodes on average, and characters are slightly more connected as shown by the degree value.

\begin{table*}[ht]
\centering
\small
\begin{tabular}{lcccccccccc}
\toprule
Split & {Block} & {Char} & $M_c$ & $V^\mathcal{S}$ & $V^\mathcal{C}$ & {Seg/Block} & $E^\mathcal{C}$ & $E^\mathcal{S}$ & {Degree$^\mathcal{C}$} & {Degree$^\mathcal{S}$} \\
\midrule
1500 Tokens & 42.3 & 17.8 & 74.2 & 250.2 & 717.3 & 17.8 & 256.7 & 492.3 & 6.5 & 1.3\\
1 Chapter & 23.4 & 18.6 & 79.5 & 167.3 & 746.2 & 42.1 & 220.4 & 529.0 & 10.5 & 1.4 \\
\hline
\end{tabular}
\caption{Summary statistics of DHCNs extracted from all twenty-thousand books.}
\label{tab:desc_stats}
\end{table*}



\section{GraphLit Configuration and Hyperparameters}
\label{app:training_details}

In this section, we report the hyperparameters used during training, as well as the general model configuration of GraphLit.
Qwen3-4B-Embeddings has a hidden dimension of $2560$, from which we project both segment and character representations to the hidden dimension of HGT.
The Transformer uses $4$ Llama decoder layers \cite{grattafiori2024llama3herdmodels} and operates directly on the projected segment representations. Each layer has $4$ attention heads and an MLP factor of $2$ for the output of self-attention.
In all GraphLit experiments, we set the hidden dimension of HGT to $256$ and use $L=3$ HGT layers, where each layer is composed of $8$ attention heads.
The HGT decoder used for attribute restoration is a single-layer HGT decoder with $8$ attention heads.
All pooling layers are composed of $8$ attention heads and project the final representations to a latent dimension of $256$.
GraphLit has a total of 8,206,688 parameters.
For the prediction heads, we employ MLP layers with GELU activations. For link prediction, we use a $2$-layer MLP with an intermediate dimension of $1024$, and for graph-level prediction, we use a $3$-layer MLP with intermediate dimensions of $1024$.



\section{Character Embedding Benchmark}
\label{app:CEB}

The Character Embedding Benchmark (CEB) \cite{inoue-etal-2022-learning} is a benchmark suite designed to evaluate fixed-length representations of fictional characters in novels.
It is derived from 17,275 public-domain books from Project Gutenberg.
The benchmark is intended to test whether character embeddings encode information about individual character attributes, local narrative context, and book-level metadata.
We describe below the different tasks presented in CEB.


\textbf{Gender} is a binary classification task in which the model predicts whether a character is male or female. Gold labels are assigned heuristically from the majority of male or female pronouns coreferent with the character. A character is labelled male if male pronouns outnumber female pronouns by at least 10\%, and female under the converse condition. The dataset originally contains 5,000 examples from which we could recover 4,216 data points.

\textbf{Role} asks the model to identify a character’s role, such as \textit{schoolmaster} or \textit{aristocrat}, from four candidate roles.
Gold roles are extracted from two reference books on English literature, with dependency parsing used to retain head nouns. This task originally contains 484 examples, from which we recover 418 data points.

\textbf{Protagonist} is a binary task that asks whether a character is the protagonist of a book. The most frequently mentioned character in a book is used as an approximate gold-standard protagonist. This task originally contains 5,000 examples, from which we recover 4,315 data points.

\textbf{Identity} is a binary pairwise task in which the model determines whether two characters from different books are the same entity. Positive examples are constructed from characters that share the same full name and author. This task originally contains 5,000 examples, from which we recover 4000 examples.

\textbf{Speaker} presents an explicit quotation of at least 50 words and asks the model to identify which character uttered it among 4 candidates. This task originally contains 2,879 examples, from which we recover 1490 data points.

\textbf{Character Identification} presents a description of a character where all mentions of the character are replaced by a [MASK] token, and asks the model to identify which characters among 4 candidates it corresponds to. We use descriptions from LiSCU \cite{brahman-etal-2021-characters-tell} to which we could map original characters to our DHCNs characters. We build the candidate subsets by always choosing the characters that are the most mentioned in the story, which we believe can be strong distractors.
It contains 893 examples.
 
\textbf{Sentence description} presents a single sentence description of a character (\textit{e.g.} ``A simple but honest man''), in which we ensure that it does not contain any named mention of the character it refers to. The model needs to identify which characters among 4 candidates the description corresponds to. We use descriptions from BookWorm \cite{papoudakis-etal-2024-bookworm}, mapping and building candidates using the same process as above.
It contains 1,987 examples.

\textbf{Question Answering} presents a character-related question and asks the model to select the correct character answer from four candidates. The questions are drawn from NarrativeQA \cite{narrativeqa2017} and character-related annotations from \citet{angelidis-etal-2019-book}. This task originally contains 587 examples, from which we recover 345 data points.

\textbf{Author} is a binary pairwise task in which the model predicts whether two characters, drawn from different books, come from books by the same author. This task originally contains 5,000 examples, from which we recover 3,566 data points.

\textbf{Book} is a binary pairwise task in which the model predicts whether two characters come from the same book. This task originally contains 5,000 examples, from which we recover 3,543 data points.

\textbf{Genre} asks whether a character belongs to a book with a given genre label. Since Project Gutenberg books may have multiple subject labels, the benchmark defines 11 binary classification tasks from frequent Project Gutenberg metadata subjects: 19th century, adventure stories, detective and mystery stories, fiction, historical fiction, humorous stories, juvenile fiction, love stories, science fiction, short stories, and western stories. The reported score averages across these binary genre tasks. The genre task originally contains 44,152 examples, from which we recover 37,332 data points.

\begin{table*}[!t]
    \centering
    \begin{tabular}{l|ccccccc}
    \toprule
         & \# of Unique Characters & \# of Books & Mean & Std & Median & $q_{0.05}$ & $q_{0.95}$ \\ 
        \midrule
        Excluded & 7769 & 444 & 317.3 & 266.8 & 233.0 & 106.0 & 817.0 \\
        Included & 43195 & 12386 & 302.5 & 245.1 & 217.0 & 109.0 & 776.0 \\
        \bottomrule
    \end{tabular}
    \caption{ Distribution statistics of mentions in the set of retained (Included) and filtered-out (Excluded) characters from the CEB benchmark. $q_{x}$ refers to the $x$ quantile of the distribution.}
\label{tab:CEB_chars_mentions}
\end{table*}

\begin{table*}
    \centering
    \begin{tabular}{l|cccccc|}
    \toprule
        ~ & 1 & 2 & 3 & 4 & 5 & 6 \\ \midrule
        Excluded & 25\% & 18\% & 22\% & 18\% & 10\% & 4\% \\ 
        Included & 20\% & 26\% & 26\% & 17\% & 7\% & 3\% \\ \bottomrule
    \end{tabular}
    \caption{Percentage (\%) of characters per number of aliases in the set of retained (Included) and filtered-out (Excluded) characters from the CEB benchmark.}
    \label{tab:CEB_chars_aliases}
\end{table*}

\paragraph{Evaluation Setup} 
To evaluate character embeddings, CEB allows the inclusion of characters drawn from books that were seen during pre-training.
In fact, we found that GraphLit variants were trained on around 83\% of books included in CEB, leaving 17\% of books completely unseen at testing time.
In contrast, our reimplementation of \citet{inoue-etal-2022-learning}'s Hybrid CN uses the entire corpus (the 20,413 books from which we extracted DHCNs), which offers a small advantage over GraphLit.
However, we still found GraphLit to be largely superior to the Hybrid CN approach.
For evaluation, we use 5-fold cross-validation for tasks with more than 2000 instances, and 10-fold cross-validation for tasks with fewer instances.
For binary tasks, we train a single linear layer on top of frozen embeddings.
For pairwise tasks (such as \textit{book} or \textit{author}), we use the CEB setup and train a linear classifier on top of element-wise product and differences $[c_1 \odot c_2 ; |c_1 - c_2|]$.
For scoring tasks (such as \textit{role} and \textit{qa}), we project the query to the dimension of candidate embeddings with a linear layer, and build scores by computing a dot product between the two.
As all contextual tasks require text-understanding, we project textual queries with \texttt{all-MiniLM-L12-v2} as done in CEB.
We train each probe for 20 epochs with a learning rate of $0.001$, using test-accuracy at the last epoch.
We fix the cross-validation seed and train linear probes 5 times, reporting the cross-validation accuracy averaged across model seeds.

\paragraph{Hybrid Character Network}
Since the only prior hybrid character network approach of \citet{inoue-etal-2022-learning} in not available, we reimplement it, making sure to follow  implementation details described in the original work.
The reimplementation leverages our entire corpus of 20,413 books,  ensuring that characters extracted from the CNC step are comparable across GraphLit variants and our reimplementation of the Hybrid CN, thus making reported results directly comparable.
In comparison, GraphLit uses 16,367 books for training, which is approximately 20\% fewer books than Hybrid CN.

Following \citet{inoue-etal-2022-learning}, we build a large graph with 4 nodes types: 1) author nodes, 2) book nodes, 3) character nodes and 4) types dependencies (context nodes).
Authors were found using the original Project Gutenberg metadata. Then, the following edges are created: a) book-author edges, b) book-character edges, c) character-context edges which connect character to context nodes if they have a dependency relation as described in \citet{bamman-etal-2014-bayesian} (e.g. Tom - traded) and d) character-character edges by using the same process as described in our submission, with a co-occurrence window of 10 tokens.
We use Spacy to extract typed-dependency relations of characters using the sentences in which they were mentioned as context.
Note that the resulting large graph is thus static (encodes the entire book), and connects different books together through how their different characters are connected through similar typed dependencies.

The resulting network contains around 450K nodes (364K are character nodes, 61.5K are context nodes, 20.5K book nodes and 4.7K author nodes), with 11.6 million edges.
We note that our network contains fewer characters than the one in \citet{inoue-etal-2022-learning}, which is caused by the removing minor characters with less than 10 mentions from our DHCN extraction step.

We then use the official implementation of DeepWalk \cite{Perozzi_2014}, with the hyperparameters reported in \citet{inoue-etal-2022-learning}: 20 walks/node, a walk length of 50 and an embedding size of 100.
DeepWalk learns node2vec embeddings from paths, consisting of a sequence of graph nodes extracted from the large graph.
Following DeepWalk training, we simply extract embeddings of character nodes within the large graph for characters that were mapped to the CEB characters.
Finally, the resulting CEB character embeddings are evaluated through the same probing pipeline.

\begin{table*}[ht!]
\centering
\small
\renewcommand{\arraystretch}{1}
\setlength{\tabcolsep}{4.5pt}
\begin{tabular}{lccccc|ccccc|cccc}
\toprule
& \multicolumn{5}{c}{\textbf{Character}}
& \multicolumn{5}{c}{\textbf{Context} }
& \multicolumn{4}{c}{\textbf{Book}}  \\

\cmidrule(lr){2-6}
\cmidrule(lr){7-11}
\cmidrule(lr){12-15}
& gender
& role
& prot
& id
& Avg
& spk
& maskd
& desc
& QA
& Avg
& auth
& book
& genre
& Avg 
\\
Support & 4216 & 418 & 4315 & 4000 & - & 1490 & 893 & 1987 & 345 & - & 3566 & 3543 & 37332 & - \\
\midrule 
 DHCN
& \makecell[c]{95.7\\[-0.5em]{\tiny (0.0)}}
& \makecell[c]{\textbf{48.7}\\[-0.5em]{\tiny (0.4)}}
& \makecell[c]{{82.8}\\[-0.5em]{\tiny (0.2)}}
& \makecell[c]{{99.3}\\[-0.5em]{\tiny (0.0)}}
& \makecell[c]{\textbf{81.6}\\[-0.5em]{\tiny (0.1)}}
& \makecell[c]{52.9\\[-0.5em]{\tiny (0.3)}}
& \makecell[c]{67.0\\[-0.5em]{\tiny (0.4)}}
& \makecell[c]{75.5\\[-0.5em]{\tiny (0.3)}}
& \makecell[c]{{52.1}\\[-0.5em]{\tiny (0.7)}}
& \makecell[c]{61.9\\[-0.5em]{\tiny (0.4)}}
& \makecell[c]{{76.3}\\[-0.5em]{\tiny (0.1)}}
& \makecell[c]{94.6\\[-0.5em]{\tiny (0.1)}}
& \makecell[c]{79.9\\[-0.5em]{\tiny (0.2)}}
& \makecell[c]{83.6\\[-0.5em]{\tiny (0.1)}} \\
 \midrule 
\;\; $- \mathcal{L}_{global}$
& \makecell[c]{\textbf{95.9}\\[-0.5em]{\tiny (0.0)}}
& \makecell[c]{48.1\\[-0.5em]{\tiny (0.5)}}
& \makecell[c]{82.2\\[-0.5em]{\tiny (0.1)}}
& \makecell[c]{\textbf{99.5}\\[-0.5em]{\tiny (0.0)}}
& \makecell[c]{81.4\\[-0.5em]{\tiny (0.2)}}
& \makecell[c]{54.2\\[-0.5em]{\tiny (0.4)}}
& \makecell[c]{64.4\\[-0.5em]{\tiny (0.4)}}
& \makecell[c]{73.6\\[-0.5em]{\tiny (0.1)}}
& \makecell[c]{{52.1}\\[-0.5em]{\tiny (1.2)}}
& \makecell[c]{61.1\\[-0.5em]{\tiny (0.5)}}
& \makecell[c]{65.5\\[-0.5em]{\tiny (0.4)}}
& \makecell[c]{82.7\\[-0.5em]{\tiny (0.2)}}
& \makecell[c]{79.0\\[-0.5em]{\tiny (0.2)}}
& \makecell[c]{75.7\\[-0.5em]{\tiny (0.3)}} \\
\;\; - Att Pooling
& \makecell[c]{95.9\\[-0.5em]{\tiny (0.0)}}
& \makecell[c]{43.5\\[-0.5em]{\tiny (0.6)}}
& \makecell[c]{{83.2}\\[-0.5em]{\tiny (0.1)}}
& \makecell[c]{99.3\\[-0.5em]{\tiny (0.0)}}
& \makecell[c]{80.5\\[-0.5em]{\tiny (0.2)}}
& \makecell[c]{54.3\\[-0.5em]{\tiny (0.2)}}
& \makecell[c]{66.4\\[-0.5em]{\tiny (0.4)}}
& \makecell[c]{\textbf{76.1}\\[-0.5em]{\tiny (0.1)}}
& \makecell[c]{51.6\\[-0.5em]{\tiny (0.5)}}
& \makecell[c]{62.1\\[-0.5em]{\tiny (0.3)}}
& \makecell[c]{78.7\\[-0.5em]{\tiny (0.1)}}
& \makecell[c]{95.9\\[-0.5em]{\tiny (0.1)}}
& \makecell[c]{81.2\\[-0.5em]{\tiny (0.2)}}
& \makecell[c]{85.2\\[-0.5em]{\tiny (0.1)}} \\
\;\; - Transformer & \makecell[c]{{95.8}\\[-0.5em]{\tiny (0.0)}}
& \makecell[c]{42.2\\[-0.5em]{\tiny (0.2)}}
& \makecell[c]{82.7\\[-0.5em]{\tiny (0.1)}}
& \makecell[c]{98.4\\[-0.5em]{\tiny (0.0)}}
& \makecell[c]{79.8\\[-0.5em]{\tiny (0.1)}}
& \makecell[c]{54.7\\[-0.5em]{\tiny (0.2)}}
& \makecell[c]{66.2\\[-0.5em]{\tiny (0.2)}}
& \makecell[c]{73.7\\[-0.5em]{\tiny (0.1)}}
& \makecell[c]{50.5\\[-0.5em]{\tiny (0.4)}}
& \makecell[c]{61.3\\[-0.5em]{\tiny (0.2)}}
& \makecell[c]{79.1\\[-0.5em]{\tiny (0.1)}}
& \makecell[c]{95.9\\[-0.5em]{\tiny (0.1)}}
& \makecell[c]{78.8\\[-0.5em]{\tiny (0.2)}}
& \makecell[c]{84.6\\[-0.5em]{\tiny (0.1)}} \\
\;\; - $L_{ord}$ & \makecell[c]{95.7\\[-0.5em]{\tiny (0.0)}}
& \makecell[c]{47.0\\[-0.5em]{\tiny (0.5)}}
& \makecell[c]{82.7\\[-0.5em]{\tiny (0.1)}}
& \makecell[c]{99.4\\[-0.5em]{\tiny (0.0)}}
& \makecell[c]{81.2\\[-0.5em]{\tiny (0.1)}}
& \makecell[c]{51.2\\[-0.5em]{\tiny (0.5)}}
& \makecell[c]{67.3\\[-0.5em]{\tiny (0.5)}}
& \makecell[c]{75.7\\[-0.5em]{\tiny (0.1)}}
& \makecell[c]{51.1\\[-0.5em]{\tiny (0.9)}}
& \makecell[c]{61.3\\[-0.5em]{\tiny (0.5)}}
& \makecell[c]{75.4\\[-0.5em]{\tiny (0.3)}}
& \makecell[c]{94.1\\[-0.5em]{\tiny (0.2)}}
& \makecell[c]{79.9\\[-0.5em]{\tiny (0.2)}}
& \makecell[c]{83.2\\[-0.5em]{\tiny (0.2)}} \\
\;\; + PE & \makecell[c]{95.8\\[-0.5em]{\tiny (0.0)}}
& \makecell[c]{47.0\\[-0.5em]{\tiny (0.7)}}
& \makecell[c]{83.1\\[-0.5em]{\tiny (0.1)}}
& \makecell[c]{99.4\\[-0.5em]{\tiny (0.1)}}
& \makecell[c]{81.3\\[-0.5em]{\tiny (0.2)}}
& \makecell[c]{51.0\\[-0.5em]{\tiny (0.5)}}
& \makecell[c]{67.1\\[-0.5em]{\tiny (0.3)}}
& \makecell[c]{74.7\\[-0.5em]{\tiny (0.3)}}
& \makecell[c]{49.7\\[-0.5em]{\tiny (0.7)}}
& \makecell[c]{60.6\\[-0.5em]{\tiny (0.4)}}
& \makecell[c]{77.8\\[-0.5em]{\tiny (0.2)}}
& \makecell[c]{94.1\\[-0.5em]{\tiny (0.1)}}
& \makecell[c]{80.0\\[-0.5em]{\tiny (0.1)}}
& \makecell[c]{84.0\\[-0.5em]{\tiny (0.1)}} \\
\;\; + GraphSAGE & \makecell[c]{94.8\\[-0.5em]{\tiny (0.1)}}
& \makecell[c]{43.6\\[-0.5em]{\tiny (0.4)}}
& \makecell[c]{82.2\\[-0.5em]{\tiny (0.3)}}
& \makecell[c]{98.7\\[-0.5em]{\tiny (0.1)}}
& \makecell[c]{79.8\\[-0.5em]{\tiny (0.2)}}
& \makecell[c]{52.4\\[-0.5em]{\tiny (0.4)}}
& \makecell[c]{68.1\\[-0.5em]{\tiny (0.6)}}
& \makecell[c]{73.2\\[-0.5em]{\tiny (0.3)}}
& \makecell[c]{51.9\\[-0.5em]{\tiny (0.4)}}
& \makecell[c]{61.4\\[-0.5em]{\tiny (0.4)}}
& \makecell[c]{78.7\\[-0.5em]{\tiny (0.3)}}
& \makecell[c]{96.2\\[-0.5em]{\tiny (0.1)}}
& \makecell[c]{80.1\\[-0.5em]{\tiny (0.2)}}
& \makecell[c]{85.0\\[-0.5em]{\tiny (0.2)}} \\
\;\; w/ E5-Large
& \makecell[c]{95.3\\[-0.5em]{\tiny (0.0)}}
& \makecell[c]{42.6\\[-0.5em]{\tiny (0.4)}}
& \makecell[c]{\textbf{84.2}\\[-0.5em]{\tiny (0.1)}}
& \makecell[c]{98.0\\[-0.5em]{\tiny (0.1)}}
& \makecell[c]{80.1\\[-0.5em]{\tiny (0.2)}}
& \makecell[c]{54.0\\[-0.5em]{\tiny (0.3)}}
& \makecell[c]{68.8\\[-0.5em]{\tiny (0.5)}}
& \makecell[c]{74.4\\[-0.5em]{\tiny (0.2)}}
& \makecell[c]{51.7\\[-0.5em]{\tiny (0.8)}}
& \makecell[c]{62.2\\[-0.5em]{\tiny (0.4)}}
& \makecell[c]{78.4\\[-0.5em]{\tiny (0.3)}}
& \makecell[c]{95.0\\[-0.5em]{\tiny (0.1)}}
& \makecell[c]{78.9\\[-0.5em]{\tiny (0.2)}}
& \makecell[c]{84.1\\[-0.5em]{\tiny (0.2)}} \\

\midrule
\midrule\;\; + Book Emb 
& \makecell[c]{95.6\\[-0.5em]{\tiny (0.1)}}
& \makecell[c]{39.2\\[-0.5em]{\tiny (0.6)}}
& \makecell[c]{82.6\\[-0.5em]{\tiny (0.3)}}
& \makecell[c]{99.4\\[-0.5em]{\tiny (0.0)}}
& \makecell[c]{79.2\\[-0.5em]{\tiny (0.2)}}
& \makecell[c]{53.0\\[-0.5em]{\tiny (0.3)}}
& \makecell[c]{67.2\\[-0.5em]{\tiny (0.8)}}
& \makecell[c]{75.6\\[-0.5em]{\tiny (0.3)}}
& \makecell[c]{52.4\\[-0.5em]{\tiny (1.0)}}
& \makecell[c]{62.1\\[-0.5em]{\tiny (0.6)}}
& \makecell[c]{80.3\\[-0.5em]{\tiny (0.2)}}
& \makecell[c]{99.9\\[-0.5em]{\tiny (0.0)}}
& \makecell[c]{\textbf{83.8}\\[-0.5em]{\tiny (0.2)}}
& \makecell[c]{\textbf{88.0}\\[-0.5em]{\tiny (0.1)}} \\
\midrule
\;\; Chapter Blocks 
& \makecell[c]{95.9\\[-0.5em]{\tiny (0.0)}}
& \makecell[c]{43.3\\[-0.5em]{\tiny (0.9)}}
& \makecell[c]{82.8\\[-0.5em]{\tiny (0.2)}}
& \makecell[c]{98.5\\[-0.5em]{\tiny (0.0)}}
& \makecell[c]{80.1\\[-0.5em]{\tiny (0.3)}}
& \makecell[c]{\textbf{54.8}\\[-0.5em]{\tiny (0.1)}}
& \makecell[c]{\textbf{69.1}\\[-0.5em]{\tiny (0.7)}}
& \makecell[c]{75.9\\[-0.5em]{\tiny (0.2)}}
& \makecell[c]{\textbf{53.2}\\[-0.5em]{\tiny (0.6)}}
& \makecell[c]{\textbf{63.2}\\[-0.5em]{\tiny (0.4)}}
& \makecell[c]{\textbf{80.4}\\[-0.5em]{\tiny (0.2)}}
& \makecell[c]{97.0\\[-0.5em]{\tiny (0.1)}}
& \makecell[c]{81.3\\[-0.5em]{\tiny (0.2)}}
& \makecell[c]{86.2\\[-0.5em]{\tiny (0.2)}} \\
\bottomrule
\end{tabular}
\caption{Accuracy (\%) across CEB tasks, for different ablations and input changes.}
\label{tab:abl_results}
\end{table*}

\begin{table}[h]
\centering
\setlength{\tabcolsep}{3.5pt}
\begin{tabular}{l|ccc}
\toprule
\textbf{Model} & \textbf{Tau} & \textbf{Rho} & \textbf{Rouge-S} \\
\midrule
DCN       & 0.097 & 0.136 & 0.549 \\
No CN    & 0.504 & 0.648 & 0.752 \\
Static CN & \textbf{0.550} & \textbf{0.695} & \textbf{0.775} \\
DHCN      & 0.512 & 0.657 & 0.756 \\
\;\; E5-Embs  & 0.408 & 0.534 & 0.704 \\
\midrule
{\small \citet{kim-skiena-2022-chapter}} & 0.143$^\star$ &  0.595 & 0.745\\
Chapters & 0.513 & 0.645 & 0.756 \\
\bottomrule
\end{tabular}
\caption{Ordering scores on around 1700 books from the test-split of \citet{kim-skiena-2022-chapter}. we note that the reported Tau in \citet{kim-skiena-2022-chapter} was not using the standard definition of Kendall's Tau, and is thus not comparable.}
\label{tab:ordering_scores}
\end{table}

\section{Comparison of Excluded and Included Characters in our CEB Evaluation}
\label{app:CEB_chars}

Our map from DHCNs characters extracted from CNC to CEB characters includes a filtering that removes characters that were ambiguous, defined as CNC characters that can be linked to multiple CEB characters.
This filtering step might potentially be confounded with how hard it is to recognize a character, which might be in turn confounded with downstream difficulty to classify these characters.

To provide an overview of whether or not this is the case, we analyse to which extent included and excluded CEB characters are different in terms of mentions and aliases.
The underlying hypothesis is that, if we observe large distributional differences across both character sets, then one might expect that the two sets are indeed different which could bias our evaluation.
As an example, if the set of excluded characters would only contain \textit{minor} characters (rarely mentioned characters), one could expect such characters to be harder to classify since less contextual information is available to model them.

We used the original character aliases provided in the CEB benchmark and count,  for each character, the number of occurrences of each of their aliases, followed by a comparison of the mention distribution of retained vs filtered-out characters.
The results are displayed in Table~\ref{tab:CEB_chars_mentions}.
In addition, we provide the proportion of unique characters per the number of unique aliases associated with them in Table~\ref{tab:CEB_chars_aliases}.
From Table~\ref{tab:CEB_chars_mentions}, we see no large differences in the distribution of mentions between unique characters included and excluded from the CEB benchmark.
In addition, we see from Table~\ref{tab:CEB_chars_mentions} that excluded characters show a slightly higher number of characters that have only one alias and more than 4, while included ones mostly have 2 or 3 aliases.
Thus, on the the one hand, included and excluded characters seem to have similar mention distributions, excluding the \textit{minor} characters hypothesis, but have a slightly different distribution of unique aliases.

Overall, these results are to interpret with precaution: it is hard to directly correlate the number of character aliases with "character difficulty", as this is mostly an authorial signal: some authors tend to give their main characters more aliases (\textit{e.g.} in \textit{Pride and Prejudice}, Elizabeth have many aliases), while some tend to give them less aliases (in \textit{Alice's Adventures in Wonderland}, Alice is the only name used to describe her).
Thus a complete statistical analysis would be required to provide a causal conclusion of whether or not we have a strong signal that the two sets of characters significantly differ.
However, from the main statistical trends we report, we do not see drastic differences and thus cautiously assume that these two sets are mostly similar.

\section{Additional GraphLit Results}
\label{app:add_results}

\subsection{Block Ordering}

As GraphLit variants have been trained on the task of block ordering, we provide in Table~\ref{tab:ordering_scores} ordering results of the test-split containing around 1700 novels unseen during training.
In particular, these books are novels from the test split of \citet{kim-skiena-2022-chapter}, and contain between 5 to 50 chapters.
In particular, chapter ordering has been shown to be a very challenging task, which requires dedicated models, and where LLMs tend to fail drastically \cite{chen2026longbenchprorealisticcomprehensive}.
We report standard rank correlation metrics between the order predicted by GraphLit and the ground truth order, as done in \citet{kim-skiena-2022-chapter}.
These metrics include Kendall's Tau, Spearman Rho and Rouge-S.
They all evaluate to which extent the predicted order correlates with the ground truth order, and are less strict than metrics such as position accuracy.

\paragraph{GraphLit models understand narrative order} Considering how hard the task is (the average number of chunks per novel is about $40$), Kendall's Tau and Spearman Rho values indicate that GraphLit---although not being perfect---has a great understanding of how the narrative is unfolding.
In contrast, Dynamic Character Networks, although trained for ordering, completely fail at capturing narrative order, suggesting that social relationships are not sufficient to fully understand it.
Thus, DHCNs dynamic character-segment grounding allows to better encode narrative order than pure dynamic character networks.
Interestingly, DHCNs with Static CNs seem to better capture narrative order than standard DHCNs, which might suggest that the structure encoded by the former is a more adequate modelling of how stories unfold.

\paragraph{At the chapter level, GraphLit works better than dedicated models} Through its self-supervised learning framework, GraphLit achieves better results than prior models from \citet{kim-skiena-2022-chapter} dedicated to the task of chapter ordering. Among others, these models include several fine-tuned Roberta models, while GraphLit only uses fewer than 10 million parameters. However, we note that initial node attributes play an important role in the success of the task, as shown by the performance drop when using E5 embeddings.

\subsection{CEB}

We provide in Table~\ref{tab:abl_results} additional experimental results on CEB and full per-task results for the entire set of ablations described in Section~\ref{sec:ablations}.
We also provide results for two GraphLit variations: 1) we replace blocks of 1500 tokens with blocks created at the chapter level and 2) we concatenate global book embeddings with global character embeddings before applying the probes.

\paragraph{Chapter-level DHCNs improve book-level task} Interestingly, training GraphLit with blocks at the chapter level provides better results on all author tasks.
Indeed, chapter organization is likely to be an authorial stylistic factor that can help distinguish among books.
Thus, character representations from DHCNs at the chapter level seem to better encode this authorial information.

\paragraph{Book Embeddings capture Authorial patterns} When using both character and book embeddings, we see an expected large increase on book-level tasks.
Thus, book embeddings seem to better capture surface-level information such as authorial style and genre definitions.

\begin{table*}[t!]
\centering
\setlength{\tabcolsep}{4pt}
\begin{tabular}{l|c|cc|cc}
\toprule
\textbf{Model} & \textbf{Overall} & \textbf{Non-Explicit} & \textbf{Explicit} & \textbf{Anaphoric} & \textbf{Implicit} \\
\midrule
Segment+Local+Global & 0.859 {\small $\pm$ 0.017} & 0.822 {\small $\pm$ 0.012} & 0.930 {\small $\pm$ 0.031} & 0.852 {\small $\pm$ 0.020} & 0.786 {\small $\pm$ 0.047} \\
Segment+Local        & 0.848 {\small $\pm$ 0.026} & 0.813 {\small $\pm$ 0.025} & 0.921 {\small $\pm$ 0.038} & 0.834 {\small $\pm$ 0.043} & 0.785 {\small $\pm$ 0.040} \\
Segment+Global       & 0.865 {\small $\pm$ 0.025} & 0.826 {\small $\pm$ 0.038} & \textbf{0.941} {\small $\pm$ 0.032} & 0.853 {\small $\pm$ 0.032} & 0.795 {\small $\pm$ 0.052} \\
Local+Global         & \textbf{0.865} {\small $\pm$ 0.017} & \textbf{0.833} {\small $\pm$ 0.014} & 0.930 {\small $\pm$ 0.030} & \textbf{0.859} {\small $\pm$ 0.027} & \textbf{0.804} {\small $\pm$ 0.040} \\
Global               & 0.861 {\small $\pm$ 0.021} & 0.825 {\small $\pm$ 0.027} & 0.931 {\small $\pm$ 0.037} & 0.848 {\small $\pm$ 0.025} & 0.794 {\small $\pm$ 0.052} \\
Local                & 0.845 {\small $\pm$ 0.024} & 0.808 {\small $\pm$ 0.019} & 0.921 {\small $\pm$ 0.042} & 0.827 {\small $\pm$ 0.034} & 0.786 {\small $\pm$ 0.034} \\
Segment              & 0.825 {\small $\pm$ 0.034} & 0.781 {\small $\pm$ 0.036} & 0.911 {\small $\pm$ 0.051} & 0.788 {\small $\pm$ 0.048} & 0.761 {\small $\pm$ 0.068} \\
\bottomrule
\end{tabular}
\caption{Quotation Attribution accuracy by quotation type, where the amount of character information provided to the model varies. Results are reported as the average accuracy over 5 cross-validation seeds, along with standard deviation.}
\label{tab:qa_more}
\end{table*}

\begin{table*}[t!]
    \centering
    \begin{tabular}{l|cc|cc|cc}
    \toprule
         & \multicolumn{2}{c}{\textbf{DHCN}}& \multicolumn{2}{c}{\textbf{No CN}}& \multicolumn{2}{c}{\textbf{Qwen}}  \\
         \midrule
         & Param & CI & Param & CI & Param & CI \\
         \midrule
SF & -0.084 & [-0.188, 0.021] & 0.089 & [-0.006, 0.184]& -0.616 & [-0.720, -0.511] \\
Adv &  -0.056 & [-0.160, 0.048] & 0.003 & [-0.097, 0.102]& -0.395 & [-0.505, -0.285] \\
Humor &  0.057 & [-0.070, 0.184] & 0.212 & [0.088, 0.337]& -0.562 & [-0.716, -0.408] \\
Western & -0.004 & [-0.097, 0.089] & 0.056 & [-0.028, 0.141]& -0.165 & [-0.269, -0.062] \\
Hist &  -0.093 & [-0.179, -0.006] & -0.138 & [-0.218, -0.058]& -0.506 & [-0.599, -0.413] \\
Fantasy & -0.020 & [-0.190, 0.149] & -0.026 & [-0.181, 0.129]& -0.397 & [-0.575, -0.220] \\
Juvenile & 0.102 & [0.038, 0.167] & 0.143 & [0.082, 0.203]& -0.434 & [-0.504, -0.363]  \\
\midrule
$\in$ Train & -0.111 & [-0.168, -0.053] & -0.189 & [-0.241, -0.138] & -0.032 & [ -0.089      0.025] \\
\midrule
$R^2$ & \multicolumn{2}{c}{0.327} & \multicolumn{2}{c}{0.447}& \multicolumn{2}{c}{0.279} \\
         \bottomrule
    \end{tabular}
\caption{Regression results when using genre labels as regressors. We set the genre "Detective" as the source of comparison.}
\label{tab:full_circ}
\end{table*}

\subsection{Quotation Attribution}

Quotation attribution has been shown to be a challenging task, requiring dedicated pipelines \cite{vishnubhotla-etal-2023-improving}.
It has recently been shown that augmenting dedicated models with embeddings of characters provides significant gains \cite{michel-etal-2024-improving}.
We thus follow this setup by using our learned character embeddings as additional information to perform quotation attribution on the Project Dialogism Novel Corpus \cite{vishnubhotla-etal-2022-project}, which contains around 40,000 quotations drawn from 27 novels.
However, we had to exclude a book from our evaluation setup (``Howards End'' written by E. M. Forster) as it is not part of Project Gutenberg.

\paragraph{Training Setup} Compared to the setup of \citet{michel-etal-2024-improving}, we change the following: 1) we use ModernBERT rather than SpanBERT to extract quotation and mention contextual representations, 2) we do not use coreference resolution to form candidate mentions, and instead form candidate mentions by using named mentions extracted during DHCN creation, 3) we use a larger contextual window of 200 tokens before and after the target quotation, and do not replace other quotations in the context with special tokens.
The latter point is, in fact, possible thanks to ModernBERT that extends maximum sequence length to a value larger than 512 tokens.

We found that in some cases, a percentage of quotations were \textit{unanswerable} in the sense that no available candidate mentions referring to the true speaker is available in the provided context.
Thus, we apply the following modification to the BookNLP+ architecture: for any training quotation, we consider the set of all contextual mentions augmented with characters that are not present in context.
For a character $c$ which is not mentioned in context, we build mention representation by concatenating representations from a $[start]$ and $[end]$ learnable tokens.
Then, we simply concatenate the mention representation with a character embedding $\mathbf{H}_c$ to form a candidate representation that still contains character-level information.
Thus, the model can still execute predictions when it thinks that no candidate mentions in context are referring to the speaker.
Finally, we create contextualized representations for candidates by concatenating its representation with the target quotation representation (which is itself a concatenation of its start and end token embeddings).
Candidate embeddings are fed to a 2-layer MLP to provide a scalar score.  

For training, we maximize the likelihood of all candidates referring to the ground-truth speaker.
We use a batch size of 64 quotations with a learning rate of~$7\times10^{-5}$.

\paragraph{Evaluation Setup} We follow \citet{michel-etal-2024-improving} and perform 5-fold cross-validation with dedicated splits, from which we removed the book ``Howards End''.
For the UAR$_{Scene}$ baseline, we use the original code to build stylistic character embeddings from explicit quotations, and then use them as global character representations $\mathbf{H}_c$.
Since the ModernBERT baseline does not use additional character information, we train it without the modifications applied to the candidate set mentioned above.

\paragraph{Ablations} For GraphLit variants, instead of using only the available global character representation, we also include the segment representation from which the quote was taken from as well as the local character representation derived from the block the quote was taken from (for Static CN, we do not use such local character representation as they are similar to the global representation).
In case where a character was not mentioned in the block, we instead use the nearest available local character representation.
We present in Table~\ref{tab:qa_more} additional results, where we vary the amount of information provided to the ModernBERT model to assess the utility of each type of representation.
We see that the \textbf{Global} character representation accounts for most of the downstream performance, followed by the \textit{local} character representation.
Interestingly, the best performance on non-explicit quotations is achieved when using both local and global embeddings, while the best performance on explicit quotations is achieved when combining the segment and global representations.
These results suggest that although the global representation is more informative overall, segment and local character representations encode different information types that can be leveraged to solve different kinds of tasks.

\section{Social Grounding in GraphLit}
\label{app:social_grounding}

From the CEB evaluation of Section~\ref{sec:results}, our main findings revealed that GraphLit's largest performance driver was the grounding of characters within their dynamic textual context, while retaining character relationships  only provides task-dependent improvements.

\subsection{Austen-Alike Benchmark}

Given the surprising nature of this result with respect to standard character network theory, we further investigate the role of \textit{character-character} edges through the Austen-Alike benchmark \cite{yang-anderson-2024-evaluating}.
This benchmark forms a good test-bed for such an evaluation as it features the following 3 computational similarity tasks curated by experts, on various novels written by Jane Austen:

\begin{itemize}
    \item A \textbf{Social Benchmark}: which looks at how often characters with similar social status (like gender and marital status) are grouped together.
\item A \textbf{Narrative Role Benchmark}: same but with narrative roles (hero, parent).
\item An \textbf{Expert Benchmark}: which looks at the correlation between cosine similarity of two character representations and the number of times experts describe the two characters as similar, on any type of features.
\end{itemize}

\begin{figure*}[htbp]
    \centering
        \includegraphics[width=\textwidth]{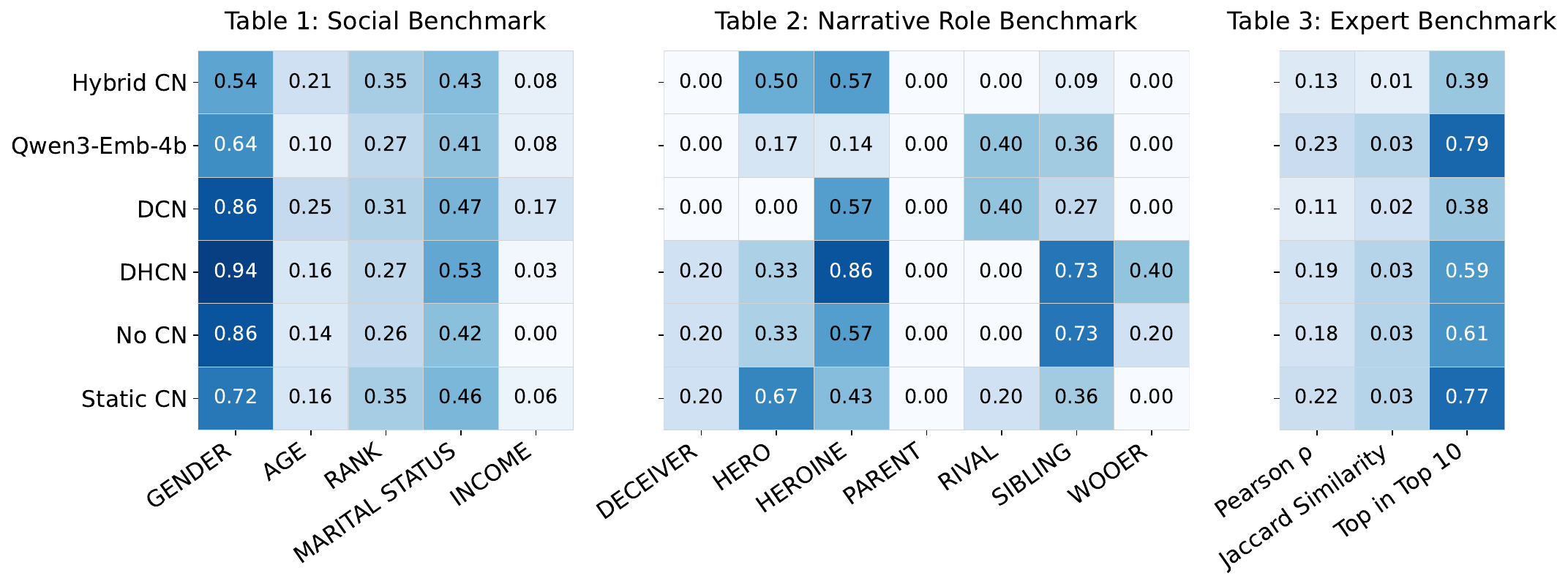}
    \caption{\textit{Left:} Average occurrence of most similar characters in the same social group by character representation. Characters from the same novel are excluded. \textit{Middle:} average occurrence of most similar character in same narrative role group by character representation. Characters from the same novel are excluded. \textit{Right:} Expert Benchmark from Austen-Alike: measures of alignment between expert pairing counts and computational similarity. Characters from  the same novel are included.}
    \label{fig:austen_alike}
\end{figure*}

The Social and Narrative Role benchmarks look at how many times a character is paired with another character that shares  similar features.
This pairing is done by associating characters that have maximal cosine similarity between their representation, across characters not present in the same novel (the pairing is done across all characters for the expert benchmark).
We run the official benchmark code for all Jane Austen novels with all GraphLit variants and baselines considered in Section~\ref{sec:results} and report the results in Figure~\ref{fig:austen_alike}.
We discuss our main findings below.

\paragraph{DCN embeddings contain social signal,} this serves as a sanity check that pure character-network features incorporate social signal that helps on the social and role benchmarks.
Indeed, DCN shows better performance than others on some social statuses, especially on Age and Income social status.
As for the role task, it performs better than Semantic embeddings on the Heroine role.

\paragraph{DHCN embeddings are systematically slightly better or similar to No CN embeddings}
On the social benchmark, they better cluster characters within their gender and marital status, which is central in Austen’s plots as her novels typically center around courtship.
On the Role benchmark, they perform similarly, except on the Heroine role, which is also an important feature of Austen's novels.
Besides, they correlate similarly with expert pairings (Table 3.). Thus, modeling explicit character-character edges seems to help in some socially grounded categories, but much of the signal already comes from character-segment grounding, which reinforces the main findings of the paper.

\paragraph{Semantic embeddings show less clustering than DHCN variants} on  social status, except for  income status, and all roles except for the rival role.
We see a particularly large difference on Gender social status and Heroine role which are important in Austen's novels.

\paragraph{DHCN improves over Hybrid CN on important Austen categories,} especially on the important Marital and Gender statuses and Heroine, Sibling, and Wooer narrative roles, but performs slightly worse on the Age, Rank and Income statuses, which are typically static attributes.
We also found it to correlate stronger with expert pairings.

\paragraph{Expert pairings appear to involve broader semantic and thematic similarity, not only social relations,} the largest correlation results are reported for the Semantic embeddings, while DCN performs  worst among tested embeddings. Interestingly, we see that the Static CN variant performs close to the Semantics embeddings. This suggests that expert similarity judgments in Austen-Alike are not exclusively relational: they also reflect broader semantic, thematic, or personality-based similarity

Overall, these results indicate that role and social status are truly social tasks, where DCN which does not use textual information still perform relatively well.
Interestingly, and as evidenced in Section~\ref{sec:results}, DHCNs without CN structure (No CN) still perform quite well, indicating that this GraphLit trained with this structure implicitly captures social relationships through character-segment grounding.
However, explicitly modeling social relationships through DHCNs seem to provide an advantage on important Austen categories such as Heroine and Marital Status.
This is consistent with our main findings: character-segment grounding is the main performance driver, as it even seems to incorporate implicit social relationship information, but explicitly modeling these relations provide task-dependent improvements.

\subsection{Qualitative Examples from Austen Novels}

Here, we focus on qualitative comparisons between full DHCNs with Hybrid CN representations.
Following the Austen-Alike benchmark, we look at pairs of maximally similar characters and check whether they seem to have similar attributes.

On the Social Status benchmark, we found, for example, that DHCN embeddings rank Jane Fa,irfax (\textit{Emma}) as the most similar character to Jane Bennet (\textit{Pride and Prejudice}), while Hybrid CN embeddings rank Sir Thomas Bertram as the most similar to Jane Bennet.

Interestingly, Jane Bennet and Jane Fairfax both share similar narrative roles in Austen's novels: they are both looking to marry, engage in romances, and serve as supporters of each novel's protagonist (Elizabeth Bennet in \textit{Pride and Prejudice} and Emma Woodhouse in \textit{Emma}).
However, they differ largely in their income: Jane Bennet is secure in the Bennet family while Jane Fairfax does not have a secure parental home.
In contrast, Sir Thomas Bertram is a wealthy landowner, associated with figures of authority, which is quite a contrastive figure compared to Jane Bennet.

We also found, that the protagonist Anne Elliot (\textit{Persuasion}) is paired with the heroine Catherine Morland (\textit{Northanger Abbey}) when using DHCNs, while paired with Sir Thomas Bertram with Inoue's embeddings, who rather serve as the Parent role.

\section{Detecting Narrative Non-Linearity}
\label{app:circuit}

As mentioned, circuitousness calculates the ratio between the distance travelled from an observed ordering of events and the minimum distance that could be travelled when taking the shortest path between events.
It is calculated as follows: suppose an ordered sequence of $T$ block embeddings $\mathbf{H}_t$.
First the euclidean distance between two succeeding blocks is computed $d_t = ||\mathbf{H}_{t+1} - \mathbf{H}_t||$, allowing to calculate \textit{Speed} $=\frac{1}{T-1}{\sum_{t=1}^{T-1}d_t}$.
Lower speed indicates that the events featured in any given sequence of text tend to be more related (where the notion of similarity is entirely defined by the representation space $\mathbf{H}_t$).
Then, \textit{Distance} is defined as the minimal distance required to cover all text blocks in the narrative, holding the first and last points fixed.
Finally, the circuitousness for a book $B$ is calculated as $circ(B) = \frac{Speed(B)}{Distance(B)}$.

Among the features we used in our regression analysis, the ratio of new characters is calculated with the following steps:
\begin{enumerate}
    \item First, for each time step $t = 0,\dots,T-1$, we find the unique characters $U_t$ present in the corresponding HCN $H_t$, and proceed similarly at time-step $t+1$ to find $U_{t+1}$ from $H_{t+1}$.
    \item Then, we compute the number of unique characters $n(U_t, U_{t+1}) = \{c, c \in U_{t+1}, c \notin U_{t}\}$ that are part of $H_{t+1}$ but not of $H_t$. The ratio is then computed as $r_t = \frac{n(U_t, U_{t+1})}{|U_{t+1}|}$.
\end{enumerate}
This ratio explicitly computes the ratio of the unique characters part of $H_{t+1}$ that were not part of $H_t$, where a high value of this ratio over all time steps indicates a constant change of new characters (an extreme value of 1 for all timesteps means that at each block, we have new characters).

Circuitousness has been shown to be lower for non-fiction books relative to fiction books, where the former are usually more linear in structure \cite{PIPER2023101793}.
Since we only focus on fiction novels, we are rather interested in looking at differences across genre when computing circuitousness with varying embeddings.
We use 8 genre tags extracted from Project Gutenberg: \textit{Science-Fiction}, \textit{Adventure Fiction}, \textit{Humorous stories}, \textit{Detective stories}, \textit{Westerns}, \textit{Historical fiction}, \textit{Fantasy} and \textit{Juvenile}.
We show in Table~\ref{tab:full_circ} the resulting regression coefficients per genre, along with their associated confidence interval, in order to interpret how genres differ in terms of linear storytelling.
We take the \textit{Detective} genre as a comparison point by dropping its genre label.

When using Qwen embeddings, we recover the analysis from \citet{PIPER2023101793} where detective stories are the most circuitous, as shown by the negative signs for all other genre.
However, the picture is drastically different when using GraphLit embeddings.
Interestingly, the differences in genre are less pronounced: only Juvenile fiction is statistically more circuitous than detective stories for both embedding types, while No CN embeddings also show higher circuitousness for humorous stories.
Besides, although the impact of being part of the training is significant for both types, it is relatively small (recall that the dependent variable is the standardized log-circuitousness, thus a value of $-0.189$ indicates that books part of the training data have, on average, a log circuitousness $-0.189$ below the mean).
Together with the fact that topological measures of social dynamics have a strong impact on non-linearity as captured by these embeddings, this suggests that DHCNs and No CN networks learn a different kind of non-linearity compared to standard semantic measures, which is not a mere training artifact.
As part of a verification, we found that the most circuitous books for both embeddings were often volumes or compilations of different works, thus often establishing new characters and settings.

\section{Computational Infrastructure}
DHCN extraction as well as GraphLit training and evaluation experiments were run on an infrastructure which contains four Nvidia Gefore GTX 1080 Ti equipped with 11.3GB GPU RAM each and two Intel(R) Xeon(R) Gold 6134 CPU @ 3.20GH with 64GB RAM each.
To build character and segment initial node attributes, along with quotation attribution experiments which involved fine-tuning ModernBERT, we instead used a single Nvidia H100, equipped with 80GB of GPU RAM.

Since DHCN extraction relies on a BERT-based Named Entity Recognition model, we heavily used GPU parallelization to achieve reasonable extraction pre-processing times.
We observed varying extraction times, ranging from 10 seconds to 2 minutes for an entire novel.
Once the DHCN were extracted, we moved them to the A100 infrastructure to encode initial node attributes in a second pass, and moved them back to the initial infrastructure for GraphLit training and evaluation.
The encoding of node attributes was done with vLLM, and resulted in the encoding of around nine millions queries (nodes) for the 20,413 books in our corpus.
On the A100 infrastructure, this took approximately 4.5 hours when using \texttt{Qwen3-4b-Embeddings}.

\end{document}